\documentclass[lettersize,journal]{IEEEtran}
\usepackage{amsmath,amssymb,amsfonts}
\usepackage{algorithmic}
\usepackage{algorithm}
\usepackage{array}
\usepackage{graphicx}
\usepackage{xcolor}
\usepackage{booktabs}
\usepackage{cite}
\usepackage{multirow}
\usepackage{threeparttable}
\usepackage[figurename=Fig., labelsep=period]{caption}  % Combined into one
\usepackage[caption=false,font=normalsize,labelfont=sf,textfont=sf]{subfig}
\usepackage{stfloats}
\usepackage{textcomp}
\usepackage{url}
\usepackage{verbatim}
\usepackage{lmodern}
\usepackage[T1]{fontenc}
\usepackage{import}
\usepackage{hyperref}
\usepackage{lmodern}
\usepackage[T1]{fontenc}
\usepackage{import}

\begin{document}

\title{Adapting SAM with Dynamic Similarity Graphs for Few-Shot Parameter-Efficient Small Dense Object Detection: A Case Study of Chickpea Pods in Field Conditions}

\author{Xintong Jiang$^{1}$, Yixue Liu$^{1}$, Mohamed Debbagh$^{1}$, Yu Tian$^{1}$, Valerio Hoyos-Villegas$^{2}$,
        Viacheslav Adamchuk$^{1}$, and Shangpeng Sun$^{1}$
\thanks{$^{1}$McGill University, Department of Bioresource Engineering, 21111 Lakeshore Road, Sainte-Anne-de-Bellevue, Quebec H9X 3V9, Canada}
\thanks{$^{2}$Michigan State University, Department of Plant, Soil and Microbial Sciences, 426 Auditorium Road, East Lansing, Michigan 48824, United States}
\thanks{Corresponding author: Shangpeng Sun (email: shangpeng.sun@mcgill.ca)}}

\maketitle

\begin{abstract}
Parameter-Efficient Fine-Tuning (PEFT) of foundation models for agricultural computer vision tasks remains challenging due to limited training data and complex field conditions. This study introduces a Dynamic Similarity-based Graph Adaptation (DSGA) module to adapt the Segment Anything Model (SAM) under extreme data constraints for precise foreground and instance segmentation of small dense objects in complex agricultural environments. Through dynamic similarity graph construction with a learnable polynomial decay-initialized weight ranking mechanism and adaptive local feature aggregation, DSGA establishes robust spatial and dynamic similarity representation with only 4.00M trainable parameters, which is 4.26\% of the original SAM. Integrating this graph-based feature adaptation with Low-Rank Adaptation (LoRA) creates a complementary optimization framework that effectively captures both local and global dependencies in image embeddings while preserving model stability and parameter efficiency. Experimental results on a challenging chickpea pod dataset demonstrated that DSGA with LoRA achieved superior performance across multiple metrics evaluated under 2, 4, 8 and 10 shots, with progressive performance gains as shot count increased. Quantitative metrics showed a 17.31\% improvement in Structure-measure and a 62.36\% gain in adaptive F-measure compared to the baseline SAM fine-tuning. Comprehensive ablation studies and visualization analyses through Grad-CAM and t-SNE validated the framework's effectiveness in feature discrimination. The proposed adaptation demonstrated practical utility for automated agricultural monitoring applications, achieving accurate pod-counting with an adjusted $\text{R}^2$ of 0.8987 for images with 10 to 120 pods under challenging field conditions.
\end{abstract}


\section{Introduction}
\noindent \IEEEPARstart{C}{omputer} vision applications in agriculture present unique challenges beyond conventional object detection domains \cite{tian2020computer, akbar2024comprehensive}. Stemming from a combination of complex environmental variations under field conditions, these challenges include severe occlusion among hundreds of small overlapping dense structures, motion blur, and perspective distortions. Additional complexities arise from the natural variability in multi-scale spatial features, maturity stages, and environmental interactions of agricultural objects, as well as limited model adaptability, computational constraints, and difficulties in building large-scale datasets that require domain expertise.

These technical barriers significantly impede crop organ-level segmentation, which constitutes a critical prerequisite for phenotypic trait analysis. Such segmentation capabilities are essential for subsequent phenotyping tasks—including morphological characterization, counting, and spatial distribution assessment of reproductive organs such as flowers, fruits, and pods—to achieve crop yield prediction and genotype selection in breeding programs. These tasks fundamentally depend on accurate detection results that remain challenging due to the aforementioned challenges, necessitating robust segmentation of complex in-field organs for proximal and remote sensing applications.

Vision foundation models, such as Contrastive Language-Image Pre-Training (CLIP) \cite{radford2021CLIP}, Segment Anything Model (SAM) \cite{kirillov2023segment} and DINOv2 \cite{oquab2023dinov2}, provide general zero-shot visual perception capabilities through knowledge retrieval from extensive pre-training on generalized datasets. While these foundation models demonstrate remarkable zero-shot generalization capabilities, their direct applications to specialized tasks have downgraded performance, especially in handling domain-specific features in complex field conditions \cite{ji2024segment, chen2023sam}. These limitations become particularly evident in few-shot scenarios, where models must adapt to new domains with minimal training data while maintaining robust performance. 

Furthermore, the utilization of large vision foundation models presents significant computational challenges, as these models are typically parameter-wise heavyweight, containing hundreds of millions to billions of parameters. Training and fully fine-tuning these models are only available with a great number of high-cost GPUs and supercomputer clusters \cite{ma2024segment}. This heavy computational resource requirement restricted the direct applications of these foundation models and resulted in a high-cost carbon footprint \cite{strubell2020energy, lacoste2019quantifying, patterson2022carbon}.

Parameter-Efficient Fine-Tuning (PEFT) has emerged as a prevalent direction for adapting foundation models to specialized domains to avoid fully fine-tuning, which minimizes the computational resources needed \cite{houlsby2019parameter, xu2023parameter}. However, existing PEFT approaches, which are mainly motivated by Natural language processing (NLP) adaptation methods, are not designed to capture the intricate spatial relationships and feature dependencies crucial for accurate vision segmentation \cite{jia2022visual}. This issue can potentially result in degraded performance when applied to complicated vision tasks. Thus, one of the major challenges lies in developing effective adaptation methods for vision foundation model that can effectively incorporate complicated domain-specific features with limited training data while maintaining the computational efficiency low enough for practical applications for field deployment.

To address these challenges, this study proposes a two-stage few-shot PEFT framework with Dynamic Similarity-based Graph Adaptation (DSGA) and Low-Rank Adaptation (LoRA). Built on top of SAM, the integrated DSGA \& LoRA module captures local and global spatial feature relationships to enhance model adaptation in limited-data scenarios, achieving robust performance with fewer than 5\% parameters of the original SAM. Comprehensive experiments demonstrate that DSGA \& LoRA achieves robust pixel-wise foreground and instance segmentation, even in cluttered environments with overlapping objects and ambiguous boundaries. The primary contributions of this work are summarized as follows:

1. Development of DSGA for vision foundation models. Integrated with LoRA, DSGA adapts SAM for complex few-shot agricultural settings. The architecture incorporates dynamic similarity adjacency graphs with learnable rank-specific weights and adaptive local feature aggregation, addressing segmentation challenges of heavy occlusion, distinct viewing angles, and multi-scale features while maintaining only 4.62\% of trainable parameters of the original SAM.

2. Introduction of a complementary two-stage adaptation framework for pixel-wise foreground and instance segmentation, achieving effective few-shot PEFT with up to 10 training images. This approach achieves State-Of-The-Art (SOTA) performance with 0.8453 S-measure for foreground segmentation and 0.7484 AP$_{50}$ with a mean Intersection over Union (IoU) of 0.8060 for instance detection.

3. Design of an efficient automated prompt generation strategy for instance-level segmentation, incorporating grid-based partitioning and adaptive distribution control mechanisms that effectively sample point prompts from dense agricultural scenes with varying object scales and occlusion patterns with minimal computational overhead.

4. Development of a composite loss function that integrates boundary-aware optimization with class-balancing components. With adaptive weight balancing, the proposed loss function addresses the significant foreground-background imbalance and enhances boundary delineation.

5. Comprehensive model interpretation through ablation studies and advanced visualization techniques of SOTA adaptation methods and their combinations, providing detailed insights into architectural component interactions and feature extraction mechanisms. The practical utility of the model is demonstrated through field deployment for chickpea pod counting.

\section{Related Work}
\noindent This section reviews recent advances in PEFT and adaptation methodologies for foundation models, with a particular focus on the SAM and its applications in agricultural computer vision.

\subsection{Vision Foundation Models}
\noindent Foundation models such as BERT \cite{devlin2019bert} and GPT \cite{radford2018improving} have revolutionized NLP \cite{bommasani2021opportunities, brown2020language}, demonstrating remarkable generalization capabilities to diverse downstream tasks. This paradigm shift has subsequently influenced computer vision, presenting distinct challenges specific to visual data feature extraction. The advent of Vision Transformer (ViT) \cite{dosovitskiy2020image} marked a significant advancement in computer vision, facilitating the development of vision foundation models. This architectural innovation that captures long-range dependencies enabled breakthroughs such as CLIP \cite{radford2021CLIP}, which employed contrastive learning between vision and language modalities to achieve substantial zero-shot capabilities. Recent developments have further expanded the capabilities of vision foundation models. DiNOv2 \cite{oquab2023dinov2} established new performance benchmarks through advanced self-supervised knowledge distillation and contrastive learning. SAM \cite{kirillov2023segment} introduced a prompt-based architecture that enables zero-shot class-agnostic segmentation, demonstrating the potential for universal visual segmentation. These advances highlight the capacity of foundation models to address complex visual understanding tasks without task-specific training data.

While these vision foundation models demonstrated unprecedented capabilities in learning generalizable representations, their scale introduces significant computational challenges. Contemporary models typically encompass hundreds of millions of parameters. For example, the SAM image encoder with a ViT-Huge backbone comprises 632M parameters. This massive parameter space, while crucial for capturing complex visual relationships, imposes substantial computational constraints during both training, fine-tuning, and inference phases.

\subsection{SAM Architecture and Capabilities}
\noindent SAM \cite{kirillov2023segment}, pre-trained on the SA-1B dataset containing 1.1 billion masks across 11 million images, constituted a significant advancement in vision foundation models. Across visual domains without task-specific fine-tuning, SAM demonstrated robust generalization capabilities.

SAM comprises three primary components: a heavyweight image encoder, a flexible prompt encoder, and a lightweight mask decoder. The image encoder incorporates a hierarchical ViT-based backbone to compute prompt-compatible image embedding. The pre-trained image encoder is available in ViT-Base, ViT-Large, and ViT-Huge, with 91M, 308M, and 632M parameters, respectively. 

The prompt encoder embeds annotator inputs as prompts, specifying segmentation targets in an image. The prompt encoder supports multiple modalities of input prompts, including point prompts for foreground and background specification, bounding boxes for region-of-interest definition, masks for refinement, and text for semantic guidance. SAM's ambiguity-aware design also supports unclear or imprecise prompts and allows promptless input. Under these cases with uncertainty, SAM generates and ranks multiple potential mask predictions.

The fast mask decoder utilizes a Transformer decoder \cite{vaswani2017attention} variant with a mask prediction head to integrate prompt information with image features, producing refined segmentation masks with confidence scores in IoU.

\subsection{Domain-Specific Adaptations of SAM}
\noindent Recent studies have extensively explored SAM's applications and adaptation across various domains. \cite{ji2024segment} comprehensively evaluated SAM on diverse real-world applications. While SAM demonstrates promising results for general object extraction, its generalizability remains constrained when processing domain-specific data. SAM requires strong prior knowledge to understand professional data and to handle complex scenes. Particularly, its efficacy diminishes in lower spatial resolution scenarios and low-contrast environments, as well as when extracting features from small and irregular objects.

In remote sensing, SAM has been adapted to efficiently segment features with regular and non-regular shapes, including roads, buildings, and land covers \cite{wang2024sampolybuild, osco2023segment, shi2025efficient, ren2024segment}. Within the agricultural domain, mask annotations and plant phenotypic traits acquisition benefit from applications of SAM \cite{carraro2023segment, xu2024automatic}. However, current implementations predominantly addressed scenarios with homogeneous backgrounds and distinct object boundaries, leaving complex agricultural conditions underexplored.

Medical imaging has seen extensive adaptation efforts on SAM. MedSAM is a foundation model for universal medical image segmentation, fully fine-tuning SAM on a large-scale dataset with more than 1.5 million mask pairs \cite{ma2024segment}. Despite MedSAM's impressive cross-modality capabilities, such large-scale fine-tuning approaches remain exceptional due to their prohibitive computational costs. To balance effectiveness and efficiency, PEFT methods have emerged as more lightweight alternatives for SAM adaptation. \cite{wu2025medical} proposed PEFT space-depth transpose adapters for 2D and 3D medical images, as well as hyper-prompting adaptation enhancing prompt-guided segmentation. Focusing on similar parameter efficiency goals but with different architectural choices, \cite{zhong2024convolution} integrated a specialized LoRA variant with lightweight convolutional operations, refining SAM's representation of local spatial context.

Despite these advances, significant challenges remain in adapting SAM to handle small objects with ambiguous edges under extremely complex field conditions. In agricultural applications, such scenarios often appear with limited training data and requires balancing parameter efficiency and computational requirements. 

\subsection{Parameter-Efficient Fine-Tuning}
\begin{figure*}[htbp]
\centering
\includegraphics[width=\textwidth]{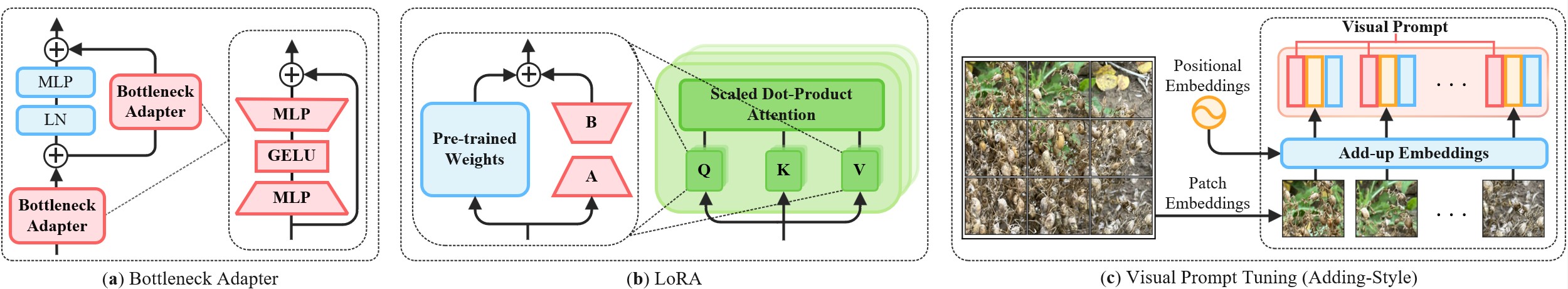}
\caption{Representative existing adaptation methods. (a) Bottleneck Adapter with dimension reduction and expansion through MLP layers, (b) LoRA adapting pre-trained weights of the Query (Q) and Value (V) projection matrices through low-rank matrices, and (c) Visual Prompt Tuning that adds learnable embeddings to the input sequence.}
\label{fig:existing_adaptation}
\end{figure*}

\noindent Fully fine-tuning a large pre-trained foundation model with parameters $\Phi_0$ creates a set of computationally intensive parameters $\Delta\Phi$ with $\vert\Delta\Phi\vert = \vert\Phi_0\vert$ for each downstream task. Partial fine-tuning paradigm limits task-specific parameter increments to $\vert\Delta\Phi\vert \ll \vert \Phi_0 \vert$, while maintaining comparable performance. Fig. \ref{fig:existing_adaptation} depicts three representative SOTA adaptation methods:

\medskip
\noindent \textbf{Bottleneck Adapters}: \cite{houlsby2019parameter} pioneered adapter-based methods that insert small-size trainable modules within transformer layers of the pre-trained model. As shown in Fig. \ref{fig:existing_adaptation}(a), the bottleneck dimension limits operational parameters for computation. These adapters are typically inserted twice, once after the multi-head attention layers and the other after the Feed-Forward Network sub-layer. AdaptFormer \cite{chen2022adaptformer} extends these methods to ViT through architectural modifications designed for visual feature processing.

\medskip
\noindent \textbf{LoRA}: \cite{hu2021lora} advanced the field by capturing a hypothesized low intrinsic rank during model weight adaptation. LoRA reparameterizes the model weight $W$ by decomposing weight updates $\Delta W$ into products of low-rank matrices as Fig. \ref{fig:existing_adaptation} (b) shows, and can be formulated as follows according to its original paper: $$h = W_0x + \Delta Wx = W_0x+\frac{\alpha}{r}BAx,$$where $W \in\mathbb{R}^{d\times k}$, and trainable parameters $B\in \mathbb{R}^{d\times r}$ and $A\in \mathbb{R}^{r\times k}$ are of a much smaller rank $r$. LoRA indirectly adapts large models with only $A$ and $B$ trainable. The scaling ${\alpha}/{r}$ reduces the necessity of retuning the hyperparameter with varying $r$. LoRA achieves significant parameter reduction, typically modifying less than 1\% of the original parameters. LoRA and its variants have been successfully applied in various computer vision tasks \cite{zhong2024convolution, zhong2024multi}. 

\medskip
\noindent \textbf{VPT}: Visual Prompt Tuning \cite{jia2022visual}, shown in Fig. \ref{fig:existing_adaptation}(c), fine-tunes ViT by inserting learnable tokens into the input embedding. These prompts achieve effective adaptation with minimal parameter overhead, despite their limited ability to capture dynamic spatial relationships essential for complex visual tasks.

While these PEFT methods provided valuable insights and contributions to model adaptation, they do not explicitly address the modeling of spatial relationships and capturing of feature interactions crucial for complex visual segmentation. Furthermore, the integration of complementary adaptation mechanisms remains an open challenge. This study addresses these limitations by introducing DSGA integrated with LoRA, offering a parameter-efficient solution to efficiently capture both global and local feature dependencies for small dense objects in complex agricultural scenes while maintaining computational feasibility. 
\section{Materials and Methods}
\noindent This section presents a PEFT two-stage framework using SAM as the foundation model, tackling the challenges in complicated agricultural segmentation tasks. Through few-shot fine-tuning, the framework effectively addresses both foreground and instance pixel-wise segmentation tasks with the proposed model components. A key component is the fused adaptation of DSGA with LoRA, which captures both global and local feature dependencies in image embeddings. The framework also incorporates an automated adaptive point distribution-aware prompt sampling strategy that generates point prompts from foreground segmentation masks, enabling efficient pixel-wise instance segmentation without manual input. Additionally, a composite loss function that compensates for class imbalance and edge precision enhancement is specifically designed for the adaptation of small dense objects.

The framework addresses the challenges of small dense plant organ detection in various agricultural applications, particularly in high-throughput plant phenotyping and precision agriculture. These segmentation challenges of multi-scale, highly occluded small objects with ambiguous boundaries in complex field backgrounds are further compounded by constraints in training data availability and computational resources, necessitating an efficient solution that maximizes performance while minimizing model parameter number. 

\subsection{Framework Overview}
\begin{figure*}[htbp]
    \centering
    \includegraphics[width=\textwidth]{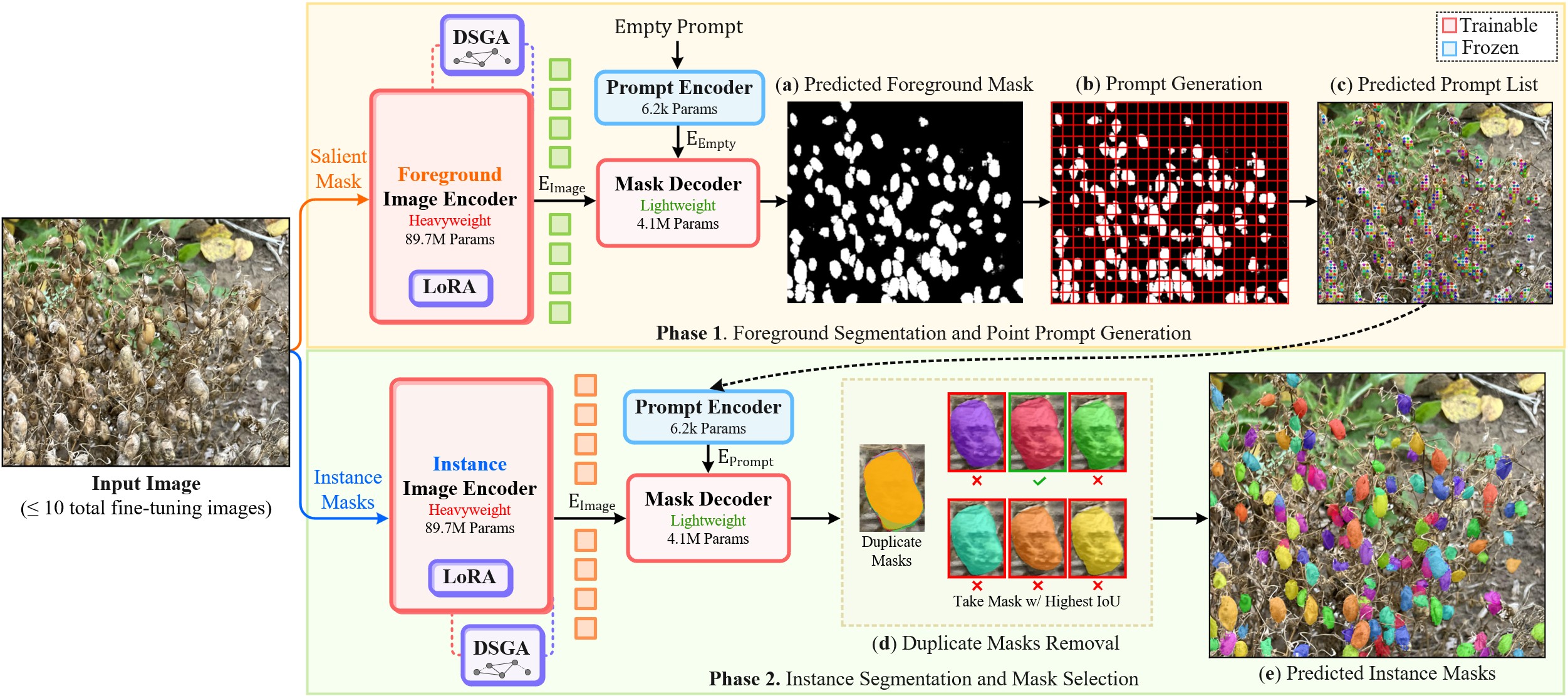}
    \caption{Architecture overview of the proposed framework. Architecture overview of the proposed PEFT framework. The two-stage process begins with promptless foreground extraction using a SAM image encoder adapted with DSGA\&LoRA. The system then automatically generates point prompts from foreground regions, which guide instance-level segmentation in the second stage. The final output retains only the instance masks with the highest predicted IoU scores, eliminating duplicates when multiple prompts target the same object.}
    \label{fig:Model_overview}
\end{figure*}

\noindent Fig. \ref{fig:Model_overview} depicts the proposed two-stage few-shot PEFT framework that sequentially performs foreground and instance segmentation. The framework employs selective parameter tuning through adaptation modules while maintaining the majority of pre-trained parameters frozen. The class-agnostic foreground image encoder in Phase 1 collectively processes all instances in an image as a single unified foreground salient mask. This image encoder selectively fine-tunes the image embedding ${E_\text{0}}$ of the input image and feeds it into the lightweight mask decoder. During this phase, the prompt encoder generates a null embedding that serves as a structural placeholder to maintain compatibility with the mask decoder's architecture, enabling promptless foreground segmentation based on learned foreground saliency features.

The transition between phases employs a grid-based strategy inspired by SAM's grid point sampling \cite{kirillov2023segment}, to automatically generate point prompts from the predicted foreground masks. The foreground mask is partitioned into uniform grid cells, where cells with high foreground occupancy generate centrally positioned point prompts. This grid-based prompt generation strategy ensures comprehensive coverage of foreground objects while maintaining computational efficiency. The generated point prompts subsequently serve as input for the prompt decoder in Phase 2, guiding the model in predicting individual instances according to each prompt.

Phase 2 conducts instance segmentation utilizing a second pre-trained SAM model, in which the image encoder takes each ground truth instance mask with a point prompt placed in its center. While the original SAM architecture supports various prompt types, implementation in this study specifically utilizes a single positive point to prompt an instance for computational efficiency. Since multiple prompts may correspond to the same instance, the framework applies a post-processing mechanism that identifies overlapping masks using IoU criteria and retains only the mask with the highest predicted confidence score. The extracted pixel-level instances enable automated quantification of chickpea pods, facilitating rapid and accurate extraction of phenotypic traits such as pod counting, size estimation, and spatial distribution analysis within each image.

The proposed PEFT architecture builds upon SAM with a pre-trained ViT-Base backbone configured with 12 transformer layers and a hidden dimension of 768, selected for its optimal balance between computational efficiency and model performance. The architecture maintains the majority of the 91M backbone parameters frozen while selectively tuning only two complementary adaptation modules in the heavyweight image encoder and the lightweight mask decoder, thereby implementing efficient fine-tuning for resource-constrained agricultural applications. 

\begin{figure*}[htbp]
    \centering
    \includegraphics[width=1\textwidth]{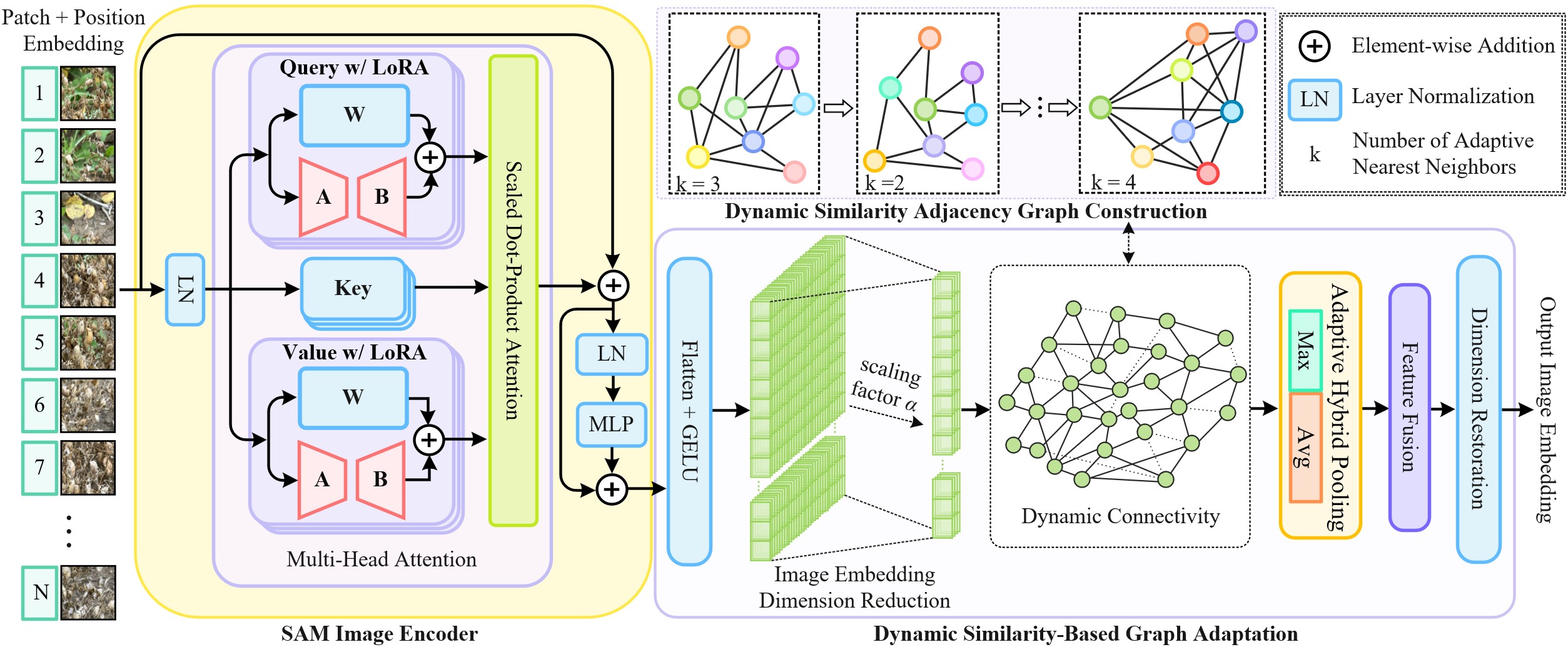}
    \caption{SAM image encoder adaptation with LoRA and DSGA. The multi-head attention section remains mostly frozen, with LoRA adaptation applied only to Query and Value projection matrices. The high-level overview of DSGA illustrates the dynamic graph construction during adaptation. While LoRA operates inside attention, DSGA modules are positioned at the terminal layers of each ViT block to process the refined image embeddings.}
    \label{fig:Image_encoder_overview}
\end{figure*}

Fig. \ref{fig:Image_encoder_overview} further illustrates the design of the image encoder with LoRA and a high-level overview of DSGA. A more detailed architecture of the DSGA component is presented in Fig. \ref{fig:DSGA_module}. LoRA operates within the multi-head attention blocks to refine feature representations through low-rank decomposed parameter updates. In the proposed model, LoRA follows the optimal structure proposed and tested by \cite{hu2021lora}, that only applies to Q and V projection matrices within the attention blocks. DSGA is positioned at the terminal layers of each ViT block to process the refined image embeddings. To address the intricate features in small dense objects, DSGA captures global dependencies through similarity-based dynamic connectivity and extracts local contextual information through adaptive feature aggregation among neighboring embedding patches. These adaptations work in concert to refine features by combining their complementary embedding modifications.

Since there are no overlapping parameters between these two modules, the total number of trainable parameters in the SAM image encoder is the number of parameters in LoRA, ${\Delta\Phi_\text{LoRA}}$, plus the number of parameters in DSGA, ${\Delta\Phi_\text{DSGA}}$, which is 4.83\% of the original image encoder's parameter $\Phi_0$ and 4.62\% of the original SAM's total parameters. This reduced parameter space provides computational advantages when fine-tuning on limited datasets, while the effectiveness of the approach is demonstrated through experimental results presented in Section 4.4.

Having established the overall network architecture, the following sections detail each key component, beginning with the Dynamic Graph-Enhanced Feature Adaptation mechanism that forms the core of the feature refinement strategy.

\subsection{Dynamic Graph-Enhanced Feature Adaptation}
\begin{figure*}[htbp]
    \centering
    \includegraphics[width=\textwidth]{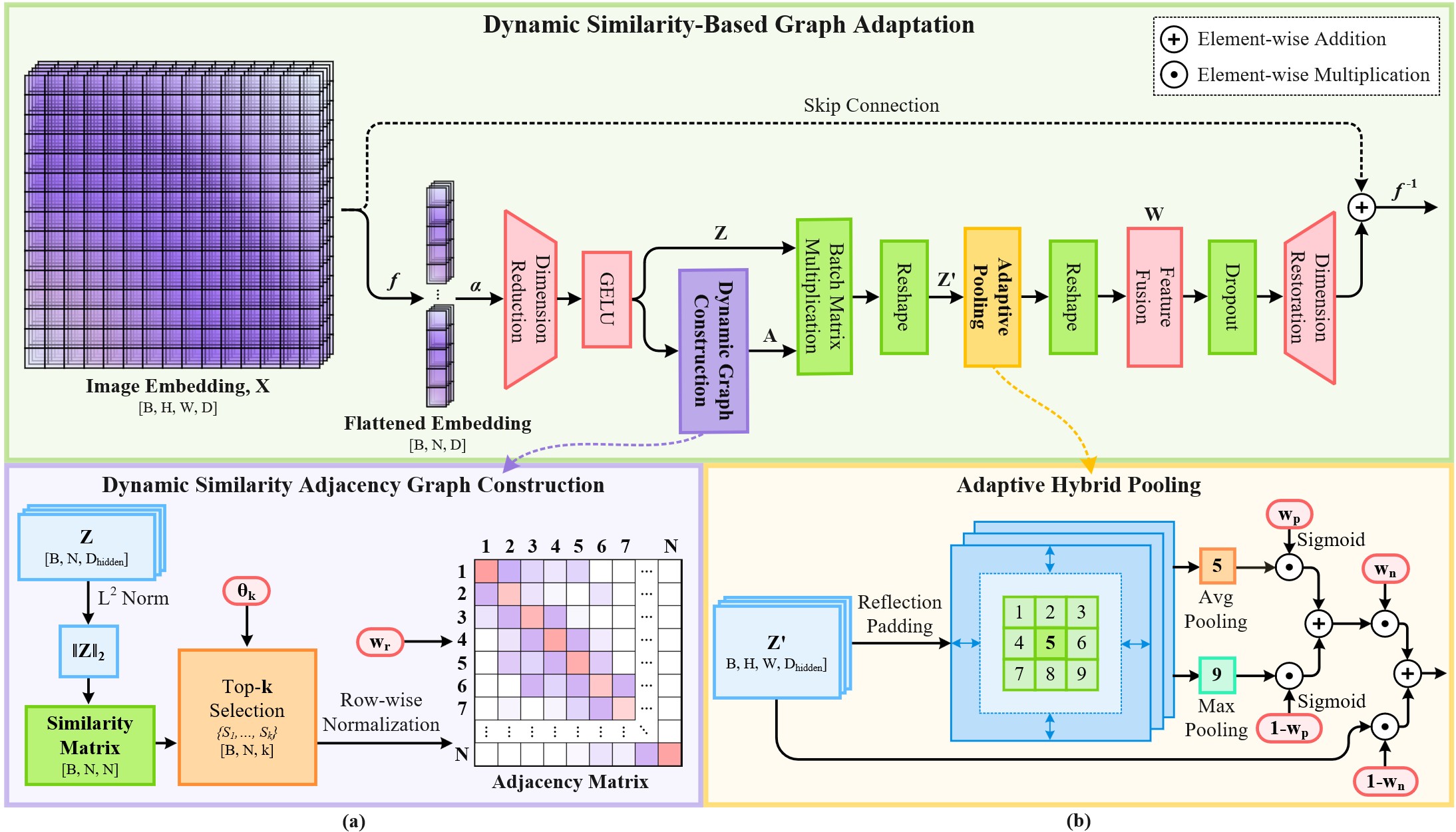}
    \caption{Architecture of the DSGA module. The bottleneck structure reduces dimensionality for DSGA to adapt features in lower-dimensional space before restoring original dimensions. The DSGA enhances feature representations through: (a) Dynamic Similarity Adjacency Graph Construction, which establishes global dependencies via $\text{L}^2$-normalized similarity computation with learnable top-k selection parameter $\theta_k$, as well as learnable rank weights $w_r$, followed by $\text{L}^1$ row-wise normalization; and (b) Adaptive Hybrid Pooling, which captures local context through weighted fusion of maximum and average operations controlled by learnable parameters $w_p$ and $w_n$. While adapting, the module preserves feature stability during adaptation with residual connections. }
    \label{fig:DSGA_module}
\end{figure*}

\noindent Fig. \ref{fig:DSGA_module} presents the design of DSGA in detail, a novel parameter-efficient adaptation mechanism designed for fine-tuning the SAM image encoder with dynamic similarity graph construction, especially for small dense object detection. Drawing inspiration from the bottleneck adapter structure \cite{houlsby2019parameter, chen2022adaptformer}, DSGA reduces and restores dimensionality via a bottleneck architecture to limit the dimension of operational parameters. With a small number of parameters inserted, this bottleneck structure preserves foundation model knowledge while efficiently incorporating domain-specific features. Distinct from original bottleneck adapters with multiple insertion points, DSGA solely integrates after the final Multi-Layer Perceptron (MLP) at the end of the encoder.

Another key component of DSGA is the dynamic similarity adjacency graph construction module, as Fig. \ref{fig:DSGA_module}(a) shows, inspired by Graph Convolutional Networks (GCN) \cite{kipf2017semi}. A dynamic sparse adjacency matrix $\mathbf{A}$ guides feature aggregation based on learnable similarity relationships among dimension-reduced embeddings. This module also introduces a learnable ranking mechanism to adjust weight distribution based on the importance of different similarity ranks. Fig. \ref{fig:DSGA_module}(b) depicts the adaptive pooling to fuse extracted features. These two modules enable the module to capture intricate spatial relationships in the embedding space, which are discussed in detail in subsequent subsections. 
 
 Given an input image embedding $\mathbf{X_{\ell}} \in \mathbb{R}^{B \times H \times W \times D}$ produced at the $\ell$-th layer of the SAM image encoder, where $B$ represents the batch size, $H \times W$ denotes the spatial dimensions, and $D$ corresponds to the embedding dimension. For implementation efficiency, DSGA first performs spatial flattening to obtain $\mathbf{X}_{\ell}'\in \mathbb{R}^{B\times N \times D}$, where $N=H \times W$ represents the number of patches. DSGA then reduces its dimension to $D_{hidden} = \lfloor \alpha D \rfloor$ with a dimension reduction ratio $\alpha$ = 0.25, which was determined experimentally to optimize the balance between computational efficiency and feature extraction capability. The DSGA operation can be formulated as:
\begin{equation}
\begin{split}
\mathbf{X}_{\ell+1} &= \underbrace{\text{DSGA}(\mathbf{X}_{\ell})}_{\text{adaptation}} + \mathbf{X}_{\ell} \\
&= f^{-1}(\mathbf{U}(\text{Drop}_{p}(\mathcal{P}(\mathbf{A}\sigma(\mathbf{D}f(\mathbf{X}_{\ell})\mathbf{W})))) + \mathbf{X}_{\ell},
\end{split}
\end{equation}

\noindent where $f: \mathbb{R}^{B \times H \times W \times D} \rightarrow \mathbb{R}^{B \times N \times D}$ is the flattening operation that reshapes the 4D tensor into a 3D tensor, $f^{-1}: \mathbb{R}^{B \times N \times D} \rightarrow \mathbb{R}^{B \times H \times W \times D}$ is the inverse operation that reshapes the 3D tensor back to the original 4D spatial dimensions, $\mathbf{D}: \mathbb{R}^{B \times N \times D} \rightarrow \mathbb{R}^{B \times N \times D_{hidden}}$ performs learnable dimension reduction and $\mathbf{U}: \mathbb{R}^{B \times N \times D_{hidden}} \rightarrow \mathbb{R}^{B \times N \times D}$ restores the original dimensionality through learnable up-projection. $\sigma$ denotes the Gaussian Error Linear Unit (GELU) activation function, $\mathbf{A}$ denotes the dynamic constructed similarity-based adjacency matrix, $\mathcal{P}$ represents the hybrid pooling operation, $\mathbf{W}$ is a learnable weight matrix for feature fusion transformation applied after graph propagation, and $\text{Drop}_{p}$ applies dropout with probability $p$. 

\subsubsection{Dynamic Similarity Adjacency Graph Construction}
\noindent The proposed dynamic graph adaptation mechanism constructs content-aware sparse adjacency matrices to model visual dependencies across feature embeddings. Unlike conventional approaches that employ static or densely connected graphs, this method adaptively captures local and global feature relationships through learned similarity metrics. The dynamic construction enables selective feature aggregation while maintaining computational efficiency through controlled sparsity. The construction process comprises three key components: similarity-based feature relationship modeling, adaptive neighborhood selection, and weighted structure aggregation. 

\medskip
\noindent\textbf{Feature Similarity Computing.}
Drawing inspiration from contrastive representation learning, the proposed approach begins with computing pairwise similarities between nodes through a temperature-controlled cosine similarity measure on normalized feature embeddings. Given an input image embedding $\mathbf{X}\in \mathbb{R}^{B\times H\times W\times D}$, it is first transformed through spatial flattening, dimension reduction operations, and GELU, resulting in an intermediate tensor $\mathbf{Z} \in \mathbb{R}^{B \times N \times D_{hidden}}$. The similarity matrix $\mathbf{S} \in \mathbb{R}^{B \times N \times N}$ is computed as:
\begin{equation}
\mathbf{S} = \tanh({\frac{{\|\mathbf{Z}\|_2}\cdot{\|\mathbf{Z}\|_2^T}}{\sqrt{D_{hidden}}}}),
\end{equation}
\noindent where $\|\mathbf{Z}\|_2$ denotes the $L^2$-normalized features. The tangent function tanh bounds the similarity values to the range [-1, 1], ensuring numerical stability in subsequent operations.

\medskip
\noindent\textbf{Dynamic Rank-specific Weight Assignment.} To effectively model the relationships between nodes in the graph and adaptively control the density of the adjacency matrix, the proposed method introduces a learnable weight mechanism with adaptive k-nearest neighbor selection. Given a maximum connectivity threshold $K_{max}$, the rank-specific learnable weights $w_r \in \mathbb{R}^{K{max}}$ are initialized using a polynomial decay pattern:

\begin{equation}
    w_r = \text{softmax}(1 - (\frac{r}{K_{\text{max}}-1})^p),
\end{equation}

\noindent where $r \in \{0,1,...,K_{\text{max}}-1\}$ represents the ranking position from the closest ($r = 0$) to the farthest ($r = K_{\text{max}}-1$) neighbor in terms of similarity, the exponent $p$ controls the polynomial decay rate. The softmax normalization ensures the weights remain positive and sum to 1, making them suitable for graph construction.
The number of neighbors k for each node is determined adaptively through:
\begin{equation}
k = \min(K_{max}, \max(1, \lfloor\text{sigmoid}(\theta_k)(K_{max}-1) + 1\rfloor)),
\end{equation}
\noindent where $\theta_k$ is initialized to $\log(\frac{K_{max}}{2})$. This adaptive mechanism employs smooth parameter transitions to facilitate stable training, allowing the network to automatically adjust its receptive field based on the spatial characteristics of the input features. The bounded range of the sigmoid function ensures $k$ remains within $[1, K_{max}]$, while the floor operation $\lfloor \cdot \rfloor$ converts the continuous output to discrete neighbor counts. 

\medskip
\noindent\textbf{Adjacency Matrix Construction.} For each node $i$, its neighborhood $\mathcal{N}_k(i)$ is constructed by selecting the top-$k$ most similar vertices based on the similarity scores:

\begin{equation}
    \mathcal{N}_k(i) = \text{topk}(S(i), k),
\end{equation}

\noindent where $S(i)$ represents the similarity scores between node $i$ and all other nodes. The adjacency matrix $\mathbf{A}$ is then constructed using the learned rank-specific weights:

\begin{equation}
    \mathbf{A}_{ij} = \begin{cases}
        w_r, & \text{if } j \in \mathcal{N}_k(i) \text{ with rank } r \\
        1, & \text{for self-connection}~(i = j) \\
        0, & \text{otherwise}
    \end{cases}
\end{equation}

To ensure balanced feature propagation, the adjacency matrix undergoes row-wise normalization, with the normalized values replacing the original entries: 

\begin{equation}
    \mathbf{A}_{ij} \leftarrow \frac{\mathbf{A}_{ij}}{\sum_{j} \mathbf{A}_{ij}}
\end{equation}

\noindent This normalization maintains consistent feature aggregation regardless of varying neighborhood sizes, which is particularly important for handling the diverse spatial patterns and feature magnitudes present in proximal sensing applications.

\subsubsection{Adaptive Local Feature Aggregation}
\noindent To enhance local spatial context modeling, an adaptive pooling mechanism is designed to dynamically fuse different pooling operations.  The tensor is reshaped back to $\mathbb{R}^{B\times H\times W\times D}$ for spatial operations. Given the learned feature maps $\mathbf{Z'} \in \mathbb{R}^{B \times H \times W \times D_{hidden}}$, the aggregation process consists of three key components:

\medskip
\noindent\textbf{Dual-path Pooling.} The mechanism employs parallel maximum and average pooling operations over $3 \times 3$ local regions. To mitigate boundary effects and preserve spatial continuity, reflective padding is applied before pooling operations, ensuring consistent feature extraction at image boundaries. This dual-path design enables the network to capture both prominent and distributed spatial patterns within local neighborhoods:
\begin{equation}
\begin{aligned}
\mathbf{P}_{\text{max}} &= \text{MaxPool}_{3\times3}(\text{Pad}_{\text{reflect}}(\mathbf{Z'})) \\
\mathbf{P}_{\text{avg}} &= \text{AvgPool}_{3\times3}(\text{Pad}_{\text{reflect}}(\mathbf{Z'}))
\end{aligned}
\end{equation}

\medskip
\noindent\textbf{Adaptive Hybrid Pooling.} The pooled features undergo learnable weighted fusion through a parameterized mechanism. A learnable parameter $w_p$ determines the relative contribution of each pooling operation. $w_p$ allows the network to automatically adjust the balance between capturing foreground features via max pooling and preserving contextual information via average pooling based on local spatial characteristics. This adaptive hybrid pooling is formulated as:
\begin{equation}
\mathbf{P} = \text{sigmoid}(w_p)\mathbf{P}_{\text{max}} + (1- \text{sigmoid}(w_p))\mathbf{P}_{\text{avg}}
\end{equation}

The fused features are integrated with the original input through a gated residual connection:
\begin{equation}
\mathbf{Z'} = (1-w_n)\mathbf{Z'} + w_n\mathbf{P},
\end{equation}
\noindent where $w_n$ is computed as $w_n = 0.5 \cdot \text{sigmoid}(w_n)$, with $w_n$ being a learnable parameter. By constraining the neighbor weight to $[0, 0.5]$ through sigmoid activation and scaling, this formulation guarantees that at least 50\% of the original features are preserved while adaptively incorporating contextual information from the pooled features. The learnable nature of $w_n$ enables the network to dynamically adjust the balance between preserving fine-grained local details and integrating a broader spatial context. This adaptive mechanism is particularly advantageous in small dense object segmentation, where spatial patterns exhibit significant variations across different geographical contexts, ranging from densely packed regions to sparse distributions.

\subsubsection{Fusion of Adaptation}
\noindent Adapter composition has proven effective for multi-task learning by combining knowledge from different adapters \cite{pfeiffer2020adapterfusion}. While primarily explored in multi-task contexts, this study applies the adapter composition to the proposed single-dataset scenarios by integrating LoRA with DSGA. These two adapters together modify the internal parameters and tuned embedding. For LoRA, the setup proposed by \cite{hu2021lora} is followed. With a decomposition rank of 8, only the Query and Value projection matrices were integrated with LoRA, leaving the Key blocks unchanged. 
Such an adapter decomposition operates at multiple levels of the network architecture, with no conflicting data flow.

\subsection{Prompt Generation}
\noindent Following the extraction of the foreground feature map, the framework transforms these refined features into spatially precise point prompts that effectively guide subsequent instance-level segmentation processes. Fig. \ref{fig:prompt_generation} illustrates the prompt generation pipeline, in which the grid-based approach converts dense foreground representations into discrete point prompts. This transformation module partitions an input foreground feature map, $\mathbf{Z''} \in \mathbb{R}^{H \times W}$, into a set of uniform grid cells, as shown in Fig. \ref{fig:prompt_generation}(a). For each cell, the module quantifies a saliency score based on foreground pixel density, identifying regions that exceed a predefined threshold for prompt generation. The prompts are positioned at the centroid of foreground pixels within identified cells. The sampling of foreground density distribution comprehensively covers foreground objects and filters out small isolated areas that are likely to be mistakenly detected as foreground. Fig. \ref{fig:prompt_generation}(b) visualizes the spatial distribution of generated point prompts derived from the foreground mask, which guides the instance mask generation in Fig. \ref{fig:prompt_generation}(c). The sampling process consists of the following components:

\begin{figure*}[htbp]
    \centering
    \includegraphics[width=\textwidth]{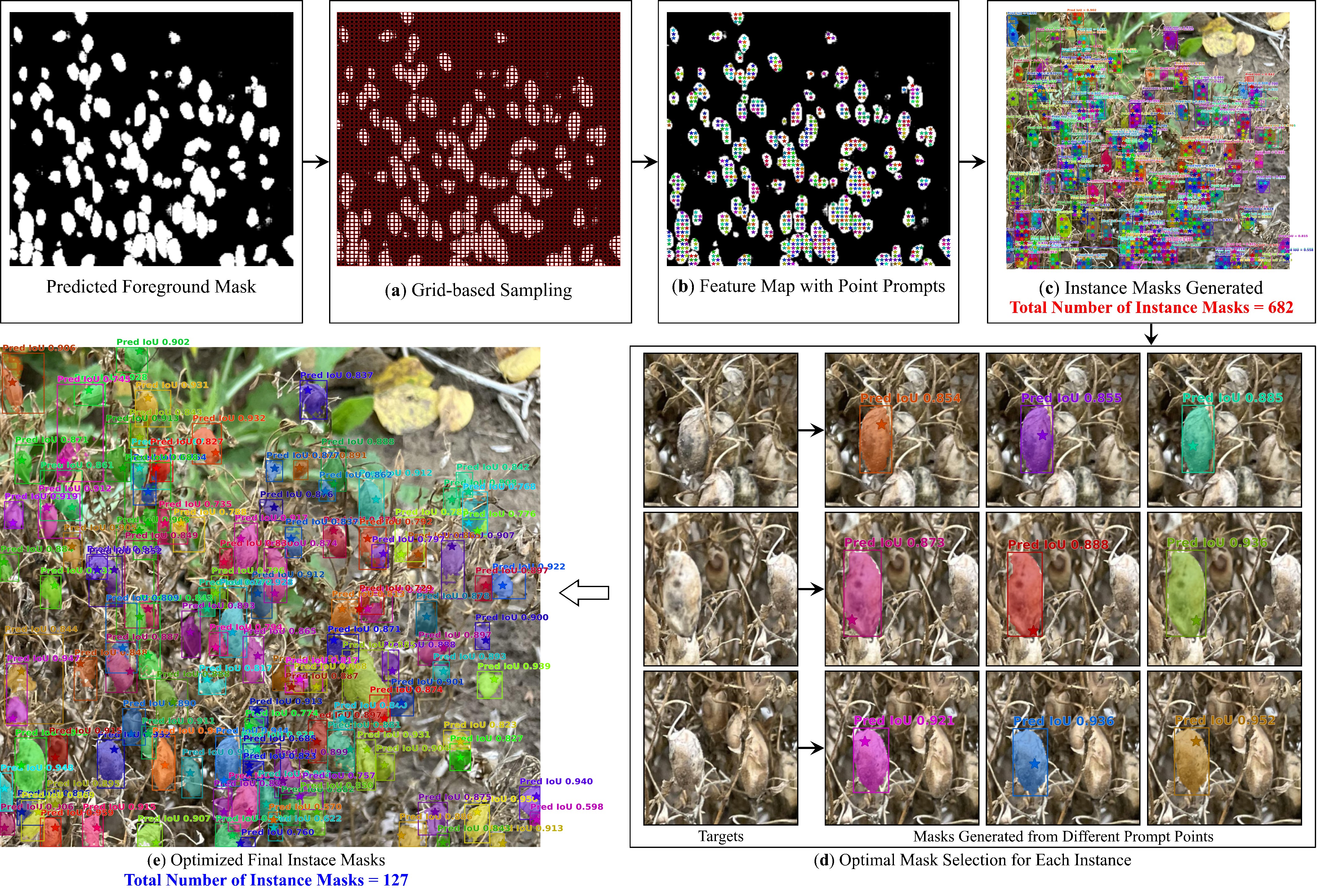}
    \caption{Prompt generation module based on grid-based sampling and instance optimization. (a) Partitioning of the predicted foreground segmentation mask into uniform grid cells with dimensions constrained by the minimum instance size to ensure comprehensive coverage. (b) Spatial distribution of automatically generated point prompts on the feature activation map, with each prompt corresponding to the region center of high semantic relevance. (c) All 682 candidate instance-level segmentation masks inferred from each generated prompt point. (d) Selection of optimal instance mask based on spatial overlapping. (e) Final optimized instance masks selected based on predicted IoUs, reducing the total instances from 678 candidates to 127 distinct masks.}
    \label{fig:prompt_generation}
\end{figure*}

\medskip
\noindent\textbf{Grid-based Partitioning.} The foreground feature map undergoes partitioning into $\lfloor\frac{H}{g}\rfloor \times \lfloor\frac{W}{g}\rfloor$ uniform grid cells of size $g \times g$  ($g=64$), leveraging computational efficiency and segmentation granularity. The grid size depends on the size of the smallest instance. For each grid cell $G_{ij}$, a saliency score $\rho_{ij}$ is computed to characterize the cell's semantic importance:
\begin{equation}
\rho_{ij} = \frac{1}{|G_{ij}|} \sum_{(x,y) \in G_{ij}} \mathbf{Z''}(x,y),
\end{equation}
\noindent where $|G_{ij}|$ represents the total pixel count within cell $G_{ij}$, and $\mathbf{Z''}(x,y)$ denotes the binary feature value $\in\{0, 1\}$ at position $(x,y)$.

\medskip
\noindent\textbf{Center Point Selection.} Within the binary feature map, pixels with value 1 serve as candidate prompt locations. For each identified grid cell, these foreground pixels contribute to determining the representative centroid:
\begin{equation}
(x_c, y_c) = (\lfloor\frac{1}{|P|}\sum_{(x,y) \in P} x\rfloor, \lfloor\frac{1}{|P|}\sum_{(x,y) \in P} y\rfloor),
\end{equation}
\noindent where $P$ denotes the set of foreground pixels within the cell. The point prompts are thus placed at the center of high semantic relevance.

\subsubsection{Adaptive Point Prompt Distribution Control}
\noindent The framework incorporates dynamic point prompt distribution control to maintain consistent performance across diverse image characteristics. When the number of generated points falls below a minimum threshold $n_{min}$, the module adaptively samples additional points from regions exhibiting the highest feature density, prioritized in descending order of selection. Similarly, each point is associated with a confidence score derived from its corresponding saliency value, facilitating weighted contribution in subsequent processing stages. To maintain computational efficiency, the total number of generated prompts is constrained by an upper bound $n_{max}$.

This adaptive approach demonstrates robust performance across varying object scales and spatial distributions. The generated point prompts constitute a compact yet comprehensive representation of foreground features, with their spatial distribution and associated confidence scores reflecting the underlying feature importance hierarchy. The generated prompts provide crucial guidance for the subsequent instance-level segmentation, leveraging the spatial information encoded in the prompts to distinguish individual instances.

\subsubsection{Instance Refinement}
 \noindent The architectural design of SAM's prompt-driven segmentation generates masks with high sensitivity to prompt placement, where minimal spatial variations in point coordinates can yield significantly different segmentation boundaries. Given the comprehensive spatial distribution of generated point prompts, candidate instance masks have overlapping predictions from proximate prompts. Fig. \ref{fig:prompt_generation}(c) depicts the complete set of instance masks generated from every point prompt. While this prompt-driven approach effectively captures subtle boundaries at varying object scales, it produces overlapping detections where multiple prompts correspond to the same underlying instance. The refinement process addresses this redundancy through the evaluation of mask overlap using IoU:
\begin{equation}
IoU_{ij} = \frac{|M_i \cap M_j|}{|M_i \cup M_j|},
\end{equation}
\noindent where $M_i$ and $M_j$ represent masks generated from prompts $p_i$ and $p_j$. When $IoU_{ij}$ exceeds the threshold $\tau_o$, only the mask with higher predict IoU is retained, effectively distilling overlapping predictions into distinct instances. The threshold $\tau_o = 0.75$ is determined through extensive empirical testing on the validation dataset to balance instance preservation with redundancy removal. Fig. \ref{fig:prompt_generation}(d) exemplifies this optimal mask selection process. The refinement step identifies and retains the optimal mask from each set of overlapping candidates that correspond to the same underlying instance. The final output instance masks are shown in Fig. \ref{fig:prompt_generation}(e), which keeps 127 instances out of the original 682 instances generated. 

\subsection{Combined Loss Function}
\noindent To address the significant imbalance of foreground and background pixel distribution and imprecise delineation of object boundaries, a composite loss function integrating three complementary components is developed. Besides the focal loss\cite{lin2017focal} and Dice Loss\cite{milletari2016v}  from the original SAM \cite{kirillov2023segment}, the proposed loss function further introduced the Boundary Loss \cite{kervadec2019boundary} . The Boundary Loss optimizes segmentation performance at the object edges, while the focal loss and Dice Loss focus on foreground and background imbalance. The proposed loss function can be formulated as follows. 

\begin{equation}
\mathcal{L}_\text{Combined}=\lambda_1\mathcal{L}_\text{Focal}+\lambda_2\mathcal{L}_\text{Dice}+\lambda_3\mathcal{L}_\text{Boundary},
\end{equation}
\noindent where $\lambda_1$, $\lambda_2$, and $\lambda_3$ are weighting factors. A variant of Dice Loss with a smoothing factor for improved numerical stability is applied. During testing, Boundary Loss significantly improves the model in identifying the boundary and thus distinguishing foreground from background.

The Exponential Moving Average (EMA) \cite{polyak1992acceleration, kendall2018multi} weight-balancing strategy is tested during training to facilitate adaptive focus without being overly sensitive to short-term fluctuations in individual loss terms. The EMA mechanism can be expressed as:
\begin{equation}
\lambda_t^{(i)} = \beta \lambda_{t-1}^{(i)} + (1-\beta)\nabla_{\lambda^{(i)}}\mathcal{L}_t,
\end{equation}
\noindent where $\lambda_t^{(i)}$ represents the weight for loss component $i$ at iteration $t$, $\beta$ denotes the momentum coefficient, and $\nabla_{\lambda^{(i)}}\mathcal{L}_t$ denotes the contribution of component $i$ to the overall loss. During training, the relative importance of different loss components can change in different stages. For example, Boundary Loss might be more critical in early training stages, while Dice Loss becomes more important later. EMA weighting allows the model to adjust focus automatically throughout the training. EMA balances the contribution of different loss terms and adapts to changes in loss magnitudes during training. Although slightly slower to adapt to changes, EMA enables smoother adaptation compared to purely dynamic weighting and provides a good balance between adaptivity and stability. EMA also reduces manual hyperparameter tuning, which is time-consuming and suboptimal. While different datasets might require different balancing of loss components, EMA accounts for dataset characteristics. Making it easier to be extended to other datasets and to potential changes in the model architecture. EMA weighting also resolved the different scales of the different loss components. While the focal loss falls into the range of 0 to 1, the range of the Boundary Loss could be much larger.  
\section{Experiments}
\subsection{Dataset}
\begin{figure*}[htbp]
    \centering
    \includegraphics[width=\textwidth]{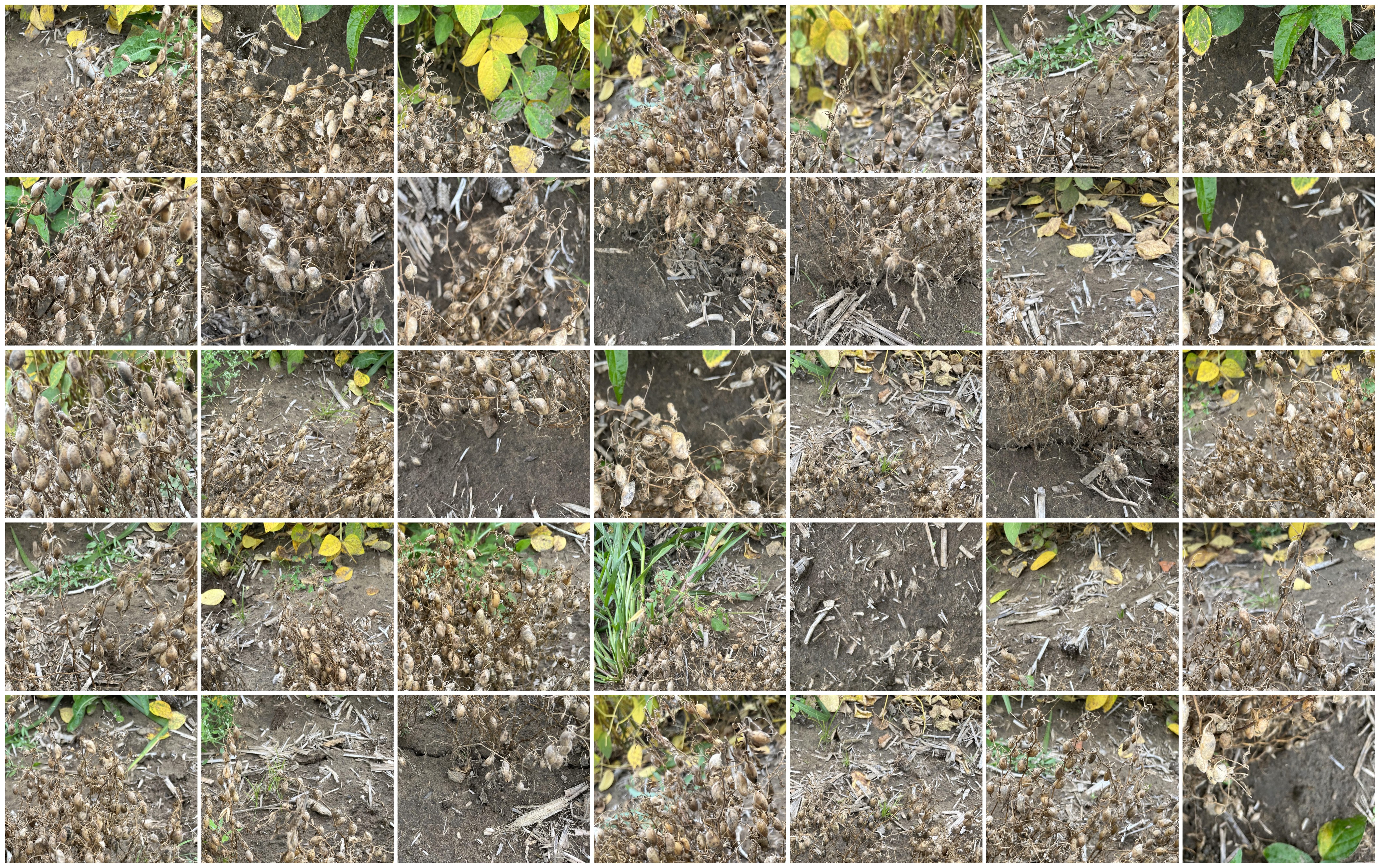}
    \caption{Illustration of the small dense chickpea pod dataset. This dataset is a demanding benchmark for few-shot foreground and instance segmentation. It presents inherent complexities in chickpea pod detection and segmentation, which is challenging even for human experts. }
    \label{fig:dataset}
\end{figure*}

\noindent This study introduces a small dense chickpea pod dataset as illustrated in Fig. \ref{fig:dataset}. Images were collected at the Emile A. Lods Agronomy Research Centre, McGill University (45°24'N, 73°56'W) during the 2022 growing season. To accurately emulate robotic viewing perspectives in agricultural environments, data acquisition was performed using a consumer-grade imaging system, which is a mobile phone equipped with a dual 12-megapixel camera system (f/1.6 wide and f/2.4 ultra-wide apertures). The device was handheld at a height of 30 cm to 50 cm above the ground, deliberately operated without image stabilization to replicate the unsteady motion typical of field robots. Images were captured while moving along 1-meter wide field alleys between chickpea rows to ensure realistic environmental representation. The dataset captures pre-harvest chickpea pods at physiological maturity stages, characterized by pod desiccation and leaf senescence. 

This real-world agricultural dataset exhibits significant complexity due to multiple factors: severe occlusions between plant structures, varying camera angles, motion-induced image degradation, depth-of-field variants, and high visual ambiguity between chickpea organs and surrounding environments. The natural variability in pod orientation and maturity stages further compounds the technical challenges. These challenges arise from both natural scene variability and acquisition conditions, complicating organ-level segmentation and making the dataset extremely hard to annotate, even for human experts.

The dataset comprises 3307 manually annotated chickpea pod instances across 60 images, with instance density varying from 9 to 118 pods per image. Following established protocols in few-shot learning research, the dataset is structured with 2, 4, 8, and 10 training images and 50 test images.

\subsection{Training Strategy}
\noindent To achieve computationally efficient adaptation, all experiments were restricted to be conducted on a single GPU. ViT-BThis study was conducted on an NVIDIA RTX 3090 GPU with 24GB memory to partially fine-tune SAM with a pre-trained ViT-Base backbone that has 91M parameters plus tested adaptation modules. All neural network linear layers were initialized using PyTorch's default initialization. For the optimizer and scheduler, this study applied AdamW optimization with betas=(0.9, 0.999) with weight decay of $1e^{-4}$  and an epsilon value of $1e^{-8}$. The learning rate schedule combined a linear warm-up for 20 epochs with a start factor of $1e^{-4}$ with cosine annealing for 180 epochs with eta min equals $1e^{-6}$. The transition between these phases was managed through PyTorch's SequentialLR scheduler with an explicit milestone at epoch 20, enabling stable initialization during warm-up while facilitating effective convergence through the annealing phase. The model resized the input images to 1024$\times$1024$\times$3 to align with SAM's architectural requirements. The experiments followed a mini-batch size of 2, which is the maximum compatible batch size. To preserve the few-shot learning paradigm's integrity, no data augmentation or additional preprocessing steps were applied beyond basic resizing and tensor conversion. This minimal intervention approach ensures the model's performance reflects its true few-shot capabilities on raw agricultural imagery.

\subsection{Evaluation Metrics}
\noindent Two different sets of metrics were applied separately to evaluate the performance of the automatic foreground segmentation and subsequent instance segmentation. For foreground segmentation, metrics for salient segmentation were applied, which are Structure-Measure ($S_\alpha$ with $\alpha=0.5$ fixed for balanced region-aware assessment) \cite{fan2017structure}, mean, maximum and adaptive F-measure ($F_\beta^\text{mean},~F_\beta^\text{max},~F_\beta^\text{adp}$, with $\beta^2 =0.3$ to emphasize precision) \cite{achanta2009frequency} and mean, maximum and adaptive E-measure ($E_\xi^\text{min},~E_\xi^\text{max},~E_\xi^\text{adp}$, where $\xi\in[0,1]$ is dynamically optimized for enhanced aligment evaluation) \cite{fan2018enhanced}, and mean absolute error (MAE, $\mathcal{M}$) \cite{perazzi2012saliency}. These metrics provide high-standard evaluations considering both region-aware and object-aware structural similarities. For instance segmentation evaluation, standard object detection metrics were utilized, which are precision, recall, F1-Score, IoU and $\text{AP}_{50}$. These metrics comprehensively assess the accuracy of instance-level segmentation results. All metrics applied in this study can be formulated as follows:

The fundamental components for evaluating segmentation accuracy are precision and recall, which form the basis for several derived metrics. Given a predicted segmentation mask $P$ and ground truth mask $G$, precision and recall are defined as

\begin{equation}
    Precision = \frac{|P \cap G|}{|P|}, \quad Recall = \frac{|P \cap G|}{|G|}
\end{equation}

These components combine to form $F_\beta$, which provides a weighted harmonic mean according to

\begin{equation}
    F_\beta = \cdot\frac{(1+\beta^2)\cdot Precision\cdot Recall}{\beta^2 \cdot Precision + Recall},
\end{equation}

\noindent where $\beta^2 = 0.3$ emphasizes precision for fine-grained segmentation tasks. With $\beta=1$, the formula computes the balanced F-measure, which is the F1-score.

$S_\alpha$ evaluates structural similarity and focuses on large-scale object structure and region-aware structural information. $S_\alpha$ builds upon these foundations to quantify structural similarity through region-aware ($S_r$) and object-aware ($S_o$) components: 

\begin{equation}
    S_\alpha = \alpha S_o + (1-\alpha)S_r,
\end{equation}

\noindent where $\alpha = 0.5$ balances regional and object-level assessments. Due to space constraints, the detailed calculation of the Structure Measure is omitted; for specifics, please refer to the original paper \cite{fan2017structure}. 

$E_\xi$ further incorporates both local and global similarities through pixel-wise alignment evaluation:

\begin{equation}
    E_\xi = \frac{1}{W\times H}\sum_{x=1}^W\sum_{y=1}^H\phi(x,y),
\end{equation}

\noindent where $\phi(x,y)$ represents the alignment score at pixel coordinates $(x,y)$.

For instance-level evaluation, the IoU metric extends the concept of spatial overlap between predicted and ground truth segmentation:

\begin{equation}
    \text{IoU} = \frac{|P\cap G|}{|P\cup G|}
\end{equation}

The average precision at 50\% IoU threshold, $AP_{50}$, aggregates precision across different recall levels while maintaining a minimum IoU requirement of 0.5:

\begin{equation}
    AP_{50} = \sum_{n}(R_n - R_{n-1})P_n,
\end{equation}

\noindent where $P_n$ and $R_n$ represent precision and recall values at the nth threshold level.

To quantify pixel-wise prediction accuracy, $\mathcal{M}$ computes the average absolute difference between predicted and ground truth masks:

\begin{equation}
    \mathcal{M} = \frac{1}{W\times H}\sum_{x=1}^W\sum_{y=1}^H|P(x,y)-G(x,y)|
\end{equation}

This integrated evaluation framework enables systematic performance assessment across different experimental configurations. The combination of pixel-wise, structural, and instance-level metrics provides a comprehensive evaluation of the segmentation model's capabilities.

\section{Results and Discussion}
\subsection{Quantitative Comparison}
\noindent Comprehensive experiments compared the proposed DSGA \& LoRA with SOTA adaptation approaches and their combinations, comprising quantitative evaluations of segmentation performance and computational efficiency. Tables~\ref{tab:salient_sota_comparsion} and~\ref{tab:instance_sota_comparison} present quantitative comparisons across adapter configurations with different numbers of fine-tuning shots, providing evaluations for foreground and instance segmentation tasks, respectively. Table\ref{tab:salient_sota_comparsion} examines model efficiency through trainable parameters in SAM image encoder and computational demands in Giga Floating-Point Operations Per second (GFLOPs). Salient segmentation metrics, $S_\alpha$, $F_\beta^\text{mean}$, $F_\beta^\text{max}$, $F_\beta^\text{adp}$, $E_\xi^\text{min}$, $E_\xi^\text{max}$, $E_\xi^\text{adp}$, and $\mathcal{M}$, evaluate foreground segmentation quality.

Table \ref{tab:instance_sota_comparison} extends the evaluation to instance-level metrics with generated prompts from foreground segmentation, focusing on precision, recall, F1-Score, $\text{AP}_{50}$, and IoU. To ensure fair comparisons, all methods were implemented on the same pre-trained SAM checkpoint with a ViT-Base backbone and follow identical training strategies. The lightweight mask decoder remained trainable across all implementations except for the original SAM baseline. The two baseline configurations for comparison are the original pre-trained SAM and a variant with its mask decoder fine-tuned on the target dataset. 

The compared SOTA models included the Bottleneck Adapter, LoRA, and the shallow and deep versions of VPT. Due to GPU memory constraints, VPT implementations in the experiments utilized an additive integration pattern instead of the prepend pattern, both proposed by \cite{jia2022visual}. Although the prepend pattern demonstrated superior performance in prior work, its computational requirements exceeded available resources. The additive pattern preserves the fundamental functionality of VPT while ensuring feasible deployment within the experimental constraints. 

The tables apply the following abbreviations for convenience. BA represents Bottleneck Adapter; BDL denotes the combination of BA, DSGA and LoRA; and BLV-S/BLV-D indicates the composition of BA, LoRA and the VPT-S or VPT-D variant.

As illustrated in Table \ref{tab:salient_sota_comparsion}, DSGA \& LoRA achieved the overall best foreground segmentation performance across all tested K-shot settings while maintaining one of the minimal parameter overheads. Instead of the double bottleneck structures of BA, the single bottleneck structure of DSGA reduces learnable parameters by approximately 50\% compared to BA in the SAM image encoder. The integrated DSGA \& LoRA demonstrated significant improvements over both baseline models. Compared to the decoder-only fine-tuning baseline, DSGA \& LoRA at K=10 shots reduces the MAE by 54.39\%, with $\mathcal{M}$ decreasing from 0.0842 to 0.0384, which showed remarkable stability with the lowest performance variance. DSGA \& LoRA also achieved 17.31\% improvement in $S_{\alpha}$, from 0.7206 to 0.8453, and pronounced improvement in adaptive measures, with 62.36\% gain $F_\beta^{\text{adp}}$ and 14.16\% gain in $E_\xi^{\text{adp}}$, indicating robust performance across varying field pod characteristics and image conditions prevalent in complex agricultural environments.

Although the few-shot capability of the proposed framework builds fundamentally upon SAM's general capabilities for feature extraction and representation, quantitative evaluation revealed that the original SAM performs inadequately on this domain-specific chickpea pod segmentation tasks. These results exhibited severely restricted performance of the original SAM across all evaluation metrics when applied directly to current complex segmentation tasks, necessitating effective domain-specific adaptation. Among all tested adaptation methods, DSGA \& LoRA showed the most exceptional adaptability, stemming from its integration of two complementary mechanisms. While LoRA alone achieved moderate performance in both segmentation tasks with 0.33M parameters, its efficient parameter updates within the attention mechanism synergistic improves the graph-based embedding fine-tuning at terminal layers via DSGA with no overlapping parameters. Other adaptation compositions, such as BDL, had performance degradation compared to DSGA \& LoRA. Although their modules do not explicitly share overlapping parameters, such degradation potentially relates to conflicting gradient updates of shared embedding spaces or information loss during the sequential embedding dimensionality reduction operations. 

Table~\ref{tab:instance_sota_comparison} demonstrates DSGA \& LoRA's robust performance in subsequent instance segmentation tasks over all tested number of shots. Compared with other methods, DSGA \& LoRA successfully detected the highest proportion of instances with the least false positive detections. With more challenging instances that the other methods fail to capture, DSGA \& LoRA preserves high IoUs for these intrinsic instances. With K=10 shots, DSGA \& LoRA achieved the highest precision of 0.7330, recall of 0.7555, F1-Score of 0.7441 and $\text{AP}_{50}$ of 0.7484. It also maintained a high average IoU of 0.8060 with more challenging instances detected. For other challenging K-shots\textless10, DSGA \& LoRA and DSGA shared most of the top metrics. The consistent outstanding performance across both foreground and instance metrics proves the DSGA \& LoRA's robustness and reliability in complex domain-specific segmentation. 

Besides the outstanding performance of the proposed module, this analysis also investigated alternative SOTA methods and their combinations, yielding several noteworthy findings. Despite proven success in other domains and substantial computational demands, VPT variants demonstrated limited effectiveness when applied solely and when integrated with other adapters on the target dataset. VPTs also exhibit performance degradation when integrated with other tested methods. For example, BLV-S and BLV-D exhibited decreased performance compared to BA \& LoRA. Similarly, degradation in $S_{\alpha}$ and $E_{\xi}$ was also observed when integrating BA \& LoRA, compared to the BA alone implementation. This observation suggested potential architectural incompatibilities between VPT's prompt-based adaptation mechanism and other tested SOTA methods. These degraded results also emphasized the importance of balanced compatibility when combining adaptation mechanisms, and exhibited the synergic adaptation of DSGA \& LoRA.

\begin{table*}[!htbp]
\centering
\captionsetup{justification=raggedright,singlelinecheck=false}
\caption{Quantitative comparisons with SOTA adaptation modules for foreground segmentation. K denotes the number of shots. \#Params denotes the number of trainable parameters in the SAM image encoder. Top three results for each K-shots are highlighted in \textcolor{red}{\textbf{red}}, \textcolor{blue}{\textbf{blue}} and \textcolor{green}{\textbf{green}}, respectively.}
\label{tab:salient_sota_comparsion}
\footnotesize
\setlength{\tabcolsep}{2pt}
\renewcommand{\arraystretch}{0.85}
    \begin{tabular}{lccccccccccc}
        \hline
        Model & K & \#Params (M) $\downarrow$ & GFLOPs $\downarrow$ &$S_{\alpha}  \uparrow$ & $F_\beta^{\text{mean}} \uparrow$ & $F_\beta^{\text{max}} \uparrow$ & $F_\beta^{\text{adp}} \uparrow$ & $E_\xi^{\text{mean}} \uparrow$ & $E_\xi^{\text{max}} \uparrow$  & $E_\xi^{\text{adp}} \uparrow$& $\mathcal{M} \downarrow$\\
        \hline
        Original SAM & N/A & N/A & N/A &0.4357& 0.0107& 0.1337& 0.0917& 0.2767& 0.5801& 0.6151& 0.1388\\
        \hline
        Decoder Only& 2& N/A & N/A & 0.6566& 0.5650& 0.6213& 0.5453& 0.8547& 0.9030& 0.8417&0.0996\\
        BA & 2& 7.10& 58.02& 0.6814& 0.5948& 0.6570& 0.5747& 0.8309& 0.8904& 0.8187&\color{green}\textbf{0.0950}\\
        LoRA& 2& {\color{red}\textbf{0.33}}& {\color{red}\textbf{3.62}}& 0.6683& 0.5590& 0.6239& 0.5373& 0.8336& 0.9064& 0.8102&0.1023\\
    
        VPT-S& 2& 6.29& 541.17& 0.6130& 0.4813& 0.5354& 0.4707& 0.7576& 0.8189& 0.7450&0.1290\\
        VPT-D& 2& 6.29& 1014.77& 0.6155& 0.4750& 0.6141& 0.4383& 0.7599& 0.8846& 0.7159&0.1349\\
        BA \& DSGA& 2& 11.10& 87.09& {\color{green}\textbf{0.7165}}& {\color{green}\textbf{0.6556}}& {\color{green}\textbf{0.6925}}& {\color{green}\textbf{0.5239}}& {\color{green}\textbf{0.8192}}& {\color{green}\textbf{0.9334}}& {\color{green}\textbf{0.8678}}& 0.2312\\
        BA \& LoRA& 2& 7.43& 61.64& 0.6566& 0.5364& 0.6113& 0.5151& 0.7964& 0.8895& 0.7672&0.1088\\
        BDL& 2& 11.43& 90.71& 0.6687& 0.5713& 0.6728& 0.5333& 0.8319& 0.9125& 0.8031&0.1002\\
        BLV-S& 2& 13.72& 602.82& 0.6494& 0.5223& 0.6740& 0.4684& 0.7915& 0.9250& 0.7348&0.1180\\
        BLV-D& 2& 13.72& 1076.41& 0.6389& 0.5100& 0.6406& 0.4621& 0.7839& 0.8973& 0.7331&0.1243\\
        DSGA& 2& {\color{blue}\textbf{4.00}}& {\color{blue}\textbf{29.07}}& {\color{blue}\textbf{0.7382}}& {\color{blue}\textbf{0.6607}}& {\color{blue}\textbf{0.7758}}& {\color{blue}\textbf{0.6090}}& {\color{blue}\textbf{0.8676}}& {\color{blue}\textbf{0.9454}}& {\color{blue}\textbf{0.8396}}& {\color{blue}\textbf{0.0750}}\\
        DSGA \& LoRA& 2& {\color{green}\textbf{4.33}}& {\color{green}\textbf{32.69}}& {\color{red}\textbf{0.7383}}& {\color{red}\textbf{0.6794}}& {\color{red}\textbf{0.7822}}& {\color{red}\textbf{0.6355}}& {\color{red}\textbf{0.8771}}& {\color{red}\textbf{0.9416}}& {\color{red}\textbf{0.8574}}& {\color{red}\textbf{0.0718}}\\
        \hline
        Decoder Only& 4& N/A & N/A & 0.6464& 0.5057& 0.6561& 0.4552& 0.7678& 0.9136& 0.7141&0.1326\\
        BA & 4& 7.10& 58.02& 0.7018& 0.6001& 0.6838& 0.5787& 0.8387& 0.9150& 0.8187&0.0866\\
        LoRA& 4& {\color{red}\textbf{0.33}}& {\color{red}\textbf{3.62}}& 0.6637& 0.5399& 0.6745& 0.4973& 0.8012& 0.9230& 0.7551&0.1229\\
        VPT-S& 4& 6.29& 541.17& 0.6518& 0.5122& 0.6623& 0.4639& 0.7926& 0.9175& 0.7396&0.1191\\
        VPT-D& 4& 6.29& 1014.77& 0.6420& 0.5020& 0.6378& 0.4592& 0.7833& 0.9000& 0.7380&0.1226\\
        BA \& DSGA& 4& 11.10& 87.09& 0.7085& 0.6040& 0.7093& 0.5645& 0.8425& 0.9284& 0.8012&0.0857\\
        BA \& LoRA& 4& 7.43& 61.64& 0.6814& 0.5559& 0.6359& 0.5302& 0.8288& 0.9161& 0.7892&0.0989\\
        BDL& 4& 11.43& 90.71& {\color{green}\textbf{0.7126}}& {\color{green}\textbf{0.6104}}& {\color{green}\textbf{0.7252}}& {\color{green}\textbf{0.5653}}& {\color{green}\textbf{0.8454}}& {\color{green}\textbf{0.9347}}& {\color{green}\textbf{0.8082}}& {\color{green}\textbf{0.0867}}\\
        BLV-S& 4& 13.72& 602.82& 0.6884& 0.5812& 0.6906& 0.5406& 0.8336& 0.9123& 0.7998&0.0939\\
        BLV-D& 4& 13.72& 1076.41& 0.6669& 0.5511& 0.6422& 0.5249& 0.8140& 0.8982& 0.7874&0.1040\\
        DSGA& 4& {\color{blue}\textbf{4.00}}& {\color{blue}\textbf{29.07}}& {\color{blue}\textbf{0.7476}}& {\color{blue}\textbf{0.6452}}& {\color{blue}\textbf{0.7714}}& {\color{blue}\textbf{0.5872}}& {\color{blue}\textbf{0.8688}}& {\color{red}\textbf{0.9455}}& {\color{blue}\textbf{0.8264}}& {\color{blue}\textbf{0.0763}}\\
        DSGA \& LoRA& 4& {\color{green}\textbf{4.33}}& {\color{green}\textbf{32.69}}& {\color{red}\textbf{0.7488}}& {\color{red}\textbf{0.6565}}& {\color{red}\textbf{0.7836}}& {\color{red}\textbf{0.6013}}& {\color{red}\textbf{0.8659}}& {\color{blue}\textbf{0.9440}}& {\color{red}\textbf{0.8299}}& {\color{red}\textbf{0.0758}}\\
        \hline
        Decoder Only& 8& N/A & N/A & 0.7090& 0.6676& 0.7430& 0.6118& 0.8666& 0.9356& 0.8715&0.0787\\
        BA & 8& 7.10& 58.02& {\color{green}\textbf{0.8099}}& {\color{green}\textbf{0.8120}}& {\color{green}\textbf{0.8523}}& {\color{green}\textbf{0.8002}}& {\color{green}\textbf{0.9341}}& {\color{green}\textbf{0.9530}}& {\color{green}\textbf{0.9405}}& {\color{green}\textbf{0.0459}}\\
        LoRA & 8& {\color{red}\textbf{0.33}}& {\color{red}\textbf{3.62}}& 0.7422& 0.6830& 0.7786& 0.6288& 0.8819& 0.9455& 0.8728&0.0709\\
        VPT-S & 8& 6.29& 541.17& 0.6827& 0.6173& 0.6953& 0.5869& 0.8489& 0.9234& 0.8786&0.0873\\
        VPT-D & 8& 6.29& 1014.77& 0.6815& 0.6379& 0.6758& 0.6273& 0.8808& 0.9157& 0.8910&0.0813\\
        BA \& DSGA& 8& 11.10& 87.09& 0.8073& 0.8032& 0.8511& 0.7881& 0.9370& 0.9569& 0.9427&0.0461\\
        BA \& LoRA& 8& 7.43& 61.64& 0.8021& 0.7819& 0.8407& 0.7614& 0.9291& 0.9564& 0.9296&0.0479\\
        BDL& 8& 11.43& 90.71& 0.8078& 0.7846& 0.8598& 0.7492& 0.9324& 0.9581& 0.9285&0.0484\\
        BLV-S& 8& 13.72& 602.82& 0.7711& 0.7745& 0.8057& 0.7676& 0.9213& 0.9404& 0.9278&0.0547\\
        BLV-D& 8& 13.72& 1076.41& 0.7437& 0.7487& 0.7903& 0.7372& 0.8884& 0.9028& 0.8906&0.0611\\
        DSGA& 8& {\color{blue}\textbf{4.00}}& {\color{blue}\textbf{29.07}}& {\color{blue}\textbf{0.8117}}& {\color{blue}\textbf{0.7883}}& {\color{blue}\textbf{0.8468}}& {\color{blue}\textbf{0.7678}}& {\color{blue}\textbf{0.9372}}& {\color{blue}\textbf{0.9581}}& {\color{blue}\textbf{0.9363}}& {\color{blue}\textbf{0.0456}}\\
        DSGA \& LoRA& 8& {\color{green}\textbf{4.33}}& {\color{green}\textbf{32.69}}& {\color{red}\textbf{0.8124}}& {\color{red}\textbf{0.8209}}& {\color{red}\textbf{0.8546}}& {\color{red}\textbf{0.8128}}& {\color{red}\textbf{0.9396}}& {\color{blue}\textbf{0.9563}}& {\color{red}\textbf{0.9460}}& {\color{red}\textbf{0.0438}}\\
        \hline
        Decoder Only& 10& N/A & N/A &0.7206& 0.6442& 0.7478& 0.5914& 0.8658& 0.9387& 0.8411& 0.0842\\
        BA & 10 & 7.10& 58.02&0.8209& 0.8275& 0.8641& 0.8172& {\color{green}\textbf{0.9418}}& {\color{green}\textbf{0.9586}}& {\color{green}\textbf{0.9496}}& 0.0436\\
        LoRA & 10 & {\color{red}\textbf{0.33}}& {\color{red}\textbf{3.62}}&0.7635& 0.7395& 0.7977& 0.7111& 0.9079& 0.9430& 0.9107& 0.0602\\
        VPT-S & 10 & 6.29& 541.17&0.7170&0.6436&0.7571&0.5946&0.8850&0.9377&0.8635&0.0796\\
        VPT-D & 10 & 6.29& 1014.77&0.7015&0.6431&0.7310&0.6029&0.8674&0.9314&0.8802&0.0789\\
        BA \& DSGA&10& 11.10& 87.09&0.8015&0.7984&0.8507&0.7808&0.9337&0.9547&0.9415&0.0498\\
        BA \& LoRA&10& 7.43& 61.64&0.8169&{\color{green}\textbf{0.8386}}&{\color{green}\textbf{0.8726}}&{\color{blue}\textbf{0.8293}}&0.9335&0.9525&0.9431&0.0450\\
        BDL&10& 11.43& 90.71&{\color{blue}\textbf{0.8352}}&{\color{blue}\textbf{0.8337}}&0.8788&{\color{green}\textbf{0.8169}}&{\color{blue}\textbf{0.9474}}&0.9633&{\color{blue}\textbf{0.9536}}&{\color{blue}\textbf{0.0412}}\\
        BLV-S&10& 13.72& 602.82&0.7885&0.8019&0.8394&0.7909&0.9262&0.9444&0.9348&0.0512\\
        BLV-D&10& 13.72& 1076.41&0.7831&0.7960&0.8346&0.7846&0.9225&0.9435&0.9269&0.0545\\
        DSGA& 10& {\color{blue}\textbf{4.00}}& {\color{blue}\textbf{29.07}}&{\color{green}\textbf{0.8305}}& 0.8165& {\color{red}\textbf{0.8867}}& 0.7779& 0.9385& {\color{blue}\textbf{0.9633}}& 0.9449& {\color{green}\textbf{0.0442}}\\
        DSGA \& LoRA&10& {\color{green}\textbf{4.33}}& {\color{green}\textbf{32.69}}&{\color{red}\textbf{0.8453}}&{\color{red}\textbf{0.8495}}&{\color{blue}\textbf{0.8821}}&{\color{red}\textbf{0.9602}}&{\color{red}\textbf{0.9518}}&{\color{red}\textbf{0.9650}}&{\color{red}\textbf{0.9602}}&{\color{red}\textbf{0.0384}}\\
        \hline
    \end{tabular}
\end{table*}

\begin{table*}[!htbp]
  \centering
  \begin{threeparttable}
  \captionsetup{justification=raggedright,singlelinecheck=false}
  \caption{Quantitative comparisons with SOTA methods for instance segmentation. K denotes the number of shots. Top three results are highlighted in \textcolor{red}{\textbf{red}}, \textcolor{blue}{\textbf{blue}} and \textcolor{green}{\textbf{green}}, respectively.}
  \label{tab:instance_sota_comparison}
  \footnotesize
  \setlength{\tabcolsep}{8pt}
  \renewcommand{\arraystretch}{0.8}
  \begin{tabular}{lcccccl}
  \hline
Model & K & Precision $\uparrow$ & Recall $\uparrow$ & F1-Score $\uparrow$ & AP$_{50}$ $\uparrow$  &IoU\tnote{*} $\uparrow$
\\
\hline
    Decoder-Only& 2& \textcolor{green}{\textbf{0.5918}}& 0.5637& 0.5774& 0.5544 &0.8057\\
    BA & 2& 0.4901& 0.6318& 0.5520& 0.6276 &0.7687\\
    LoRA & 2& 0.5488& 0.6385& \textcolor{green}{\textbf{0.5903}}& 0.6417 &\textcolor{blue}{\textbf{0.8144}}\\
    VPT-S & 2& 0.4475& 0.6598& 0.5333& \textcolor{red}{\textbf{0.6664}} &0.7915\\
    VPT-D & 2& 0.4833& 0.6127& 0.5404& 0.6260 &0.8045\\
    BA \& DSGA & 2& 0.5061& 0.5417& 0.5233& 0.5418 &0.7314\\
    BA \& LoRA & 2& 0.4327& \textcolor{blue}{\textbf{0.6770}}& 0.5280& \textcolor{blue}{\textbf{0.6654}} &0.7791\\
    BDL & 2& 0.4876& 0.5892& 0.5336& 0.5781 &0.7489\\
    BLV-S & 2& 0.4297& 0.6011& 0.5012& 0.6063 &0.7362\\
    BLV-D & 2& 0.4485& 0.5727& 0.5030& 0.5773 &0.7481\\
    DSGA & 2& \textcolor{red}{\textbf{0.6632}}& \textcolor{red}{\textbf{0.6922}}& \textcolor{red}{\textbf{0.6774}}& \textcolor{green}{\textbf{0.6609}} &\textcolor{red}{\textbf{0.8236}}\\
    DSGA \& LoRA & 2& \textcolor{blue}{\textbf{0.6481}}& \textcolor{green}{\textbf{0.6602}}& \textcolor{blue}{\textbf{0.6541}}& 0.6535 &\textcolor{green}{\textbf{0.8115}}\\
\hline
    Decoder-Only& 4& 0.4985& 0.6209& 0.5530& 0.6189 &\textcolor{red}{\textbf{0.8294}}\\
    BA & 4& 0.4949& 0.6707& 0.5695& 0.6567 &0.7769\\
    LoRA & 4& 0.2110& \textcolor{green}{\textbf{0.6793}}& 0.3216& \textcolor{green}{\textbf{0.6718}} &\textcolor{blue}{\textbf{0.8176}}\\
    VPT-S & 4& \textcolor{green}{\textbf{0.5405}}& 0.6164& \textcolor{green}{\textbf{0.5760}}& 0.6182 &0.8132\\
    VPT-D & 4& 0.4968& 0.6138& 0.5492& 0.6209 &0.7969\\
    BA \& DSGA & 4& 0.1266& 0.6673& 0.2129& 0.6605 &0.7452\\
    BA \& LoRA & 4& 0.4856& 0.6441& 0.5538& 0.6276 &0.7799\\
    BDL & 4& 0.5217& 0.6206& 0.5668& 0.6120 &0.7627\\
    BLV-S & 4& 0.4860& 0.5981& 0.5363& 0.5898 &0.7590\\
    BLV-D & 4& 0.4544& 0.5712& 0.5061& 0.5665 &0.7548\\
    DSGA & 4& \textcolor{red}{\textbf{0.6361}}& \textcolor{blue}{\textbf{0.7084}}& \textcolor{red}{\textbf{0.6703}}& \textcolor{blue}{\textbf{0.7037}} &\textcolor{green}{\textbf{0.8148}}\\
    DSGA \& LoRA & 4& \textcolor{blue}{\textbf{0.5866}}& \textcolor{red}{\textbf{0.7219}}& \textcolor{blue}{\textbf{0.6472}}& \textcolor{red}{\textbf{0.7088}} &0.8090\\
\hline
    Decoder-Only& 8& 0.6685& 0.5761& 0.6258& 0.5590 &\textcolor{blue}{\textbf{0.8241}}\\
    BA & 8& 0.6158& 0.6748& 0.6440& 0.6632 &0.7613\\
    LoRA & 8& \textcolor{green}{\textbf{0.6770}}& 0.6426& \textcolor{green}{\textbf{0.6594}}& 0.6221 &\textcolor{red}{\textbf{0.8253}}\\
    VPT-S & 8& 0.6157& 0.5331& 0.5714& 0.5254 &\textcolor{green}{\textbf{0.8153}}\\
    VPT-D & 8& 0.6502& 0.5308& 0.5845& 0.5255 &0.8031\\
    BA \& DSGA & 8& 0.4844& 0.5619& 0.5202& 0.5618 &0.7023\\
    BA \& LoRA & 8& 0.5684& \textcolor{blue}{\textbf{0.7207}}& 0.6356& \textcolor{blue}{\textbf{0.7043}} &0.7747\\
    BDL & 8& 0.5675& 0.6538& 0.6076& 0.6484 &0.7490\\
    BLV-S & 8& 0.5169& 0.5940& 0.5528& 0.5818 &0.7319\\
    BLV-D & 8& 0.5291& 0.5682& 0.5479& 0.5531 &0.7391\\
    DSGA & 8& \textcolor{red}{\textbf{0.7082}}& \textcolor{red}{\textbf{0.7293}}& \textcolor{red}{\textbf{0.7186}}& \textcolor{red}{\textbf{0.7293}} &0.8104\\
    DSGA \& LoRA & 8& \textcolor{blue}{\textbf{0.6816}}& \textcolor{green}{\textbf{0.7178}}& \textcolor{blue}{\textbf{0.6992}}& \textcolor{green}{\textbf{0.7021}} &0.7941\\
\hline
    Decoder-Only& 10 & 0.6559 & 0.6064 & 0.6301 & 0.5923  &\textcolor{red}{\textbf{0.8286}} \\
    BA & 10 & 0.6312 & 0.6617 & 0.6461 & 0.6501  &0.7755 \\
    LoRA & 10 & \textcolor{green}{\textbf{0.6794}} & 0.6464 & 0.6625 & 0.6482  &0.7832\\
    VPT-S & 10 & 0.6159 & 0.6079 & 0.6119 & 0.6073  &\textcolor{blue}{\textbf{0.8131}} \\
    VPT-D & 10 & 0.6246 & 0.5585 & 0.5897 & 0.5472  &0.7839 \\
    BA \& DSGA & 10 & 0.5297 & 0.5802 & 0.5538 & 0.5773  &0.7286 \\
    BA \& LoRA & 10 & 0.6599 & \textcolor{green}{\textbf{0.6890}} & \textcolor{green}{\textbf{0.6741}} & \textcolor{green}{\textbf{0.6784}}  &0.7795\\
    BDL & 10 & 0.5564 & 0.6120 & 0.5829 & 0.6004  &0.7390 \\
    BLV-S & 10 & 0.5641 & 0.5753 & 0.5697 & 0.5579  &0.7385 \\
    BLV-D & 10 & 0.5690 & 0.5888 & 0.5787 & 0.5848  &0.7368 \\
    DSGA & 10 & \textcolor{blue}{\textbf{0.6842}} & \textcolor{blue}{\textbf{0.7226}} & \textcolor{blue}{\textbf{0.7029}} & \textcolor{blue}{\textbf{0.7194}}  &0.8051 \\
    DSGA \& LoRA & 10 & \textcolor{red}{\textbf{0.7330}} & \textcolor{red}{\textbf{0.7555}} & \textcolor{red}{\textbf{0.7441}} & \textcolor{red}{\textbf{0.7484}}  &\textcolor{green}{\textbf{0.8060}} \\
\hline
\end{tabular}
    \begin{tablenotes}
      \small
      \item[*] {Only IoUs of detected instances were measured. Thus, IoU does not directly reflect detection capability, as challenging cases often yield lower IoU scores and can only be detected by better-performing models.}
    \end{tablenotes}
  \end{threeparttable}
\end{table*}

\subsection{Qualitative Analysis}
\begin{figure*}[htbp]
    \centering
    \includegraphics[width=\textwidth]{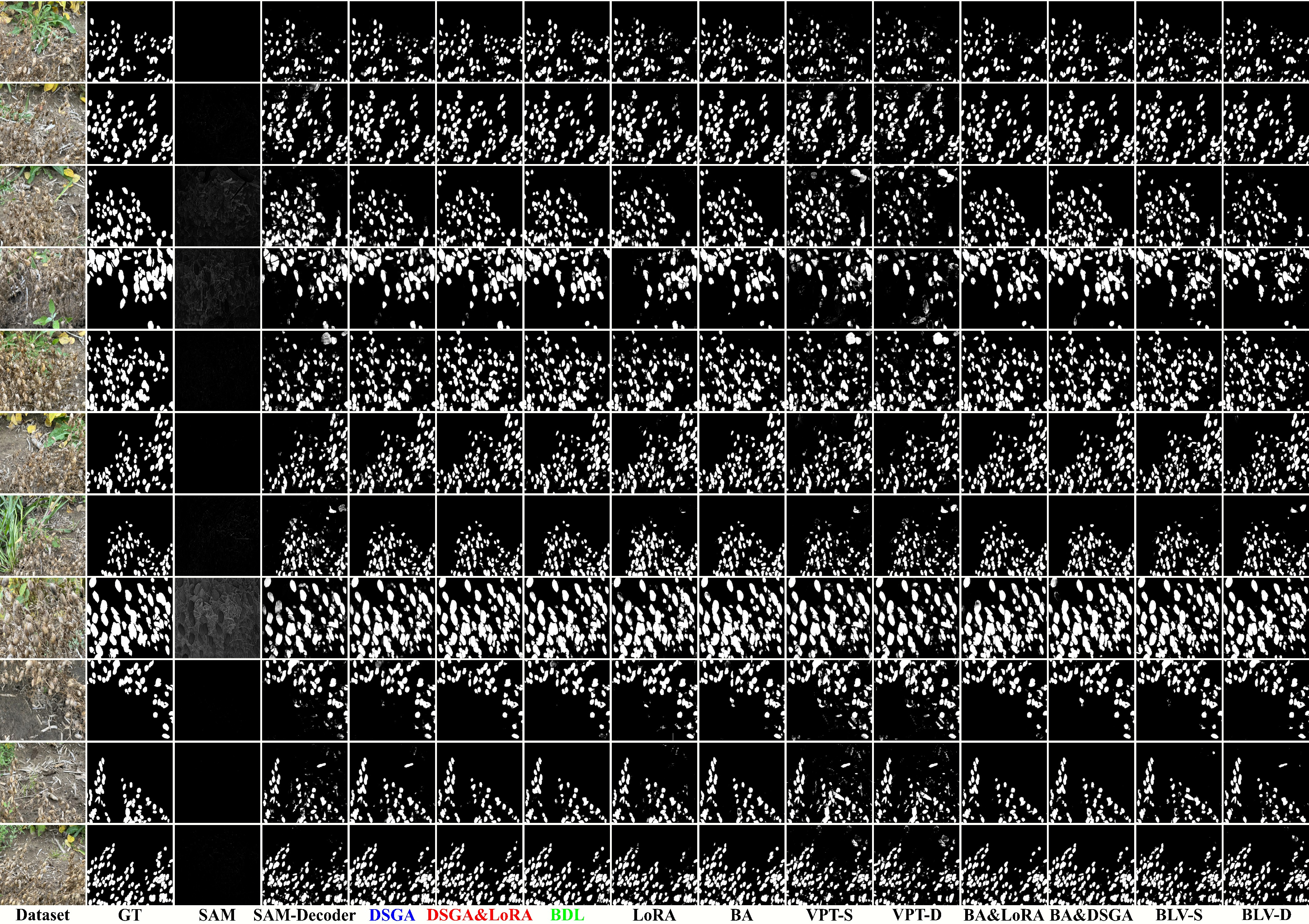}
    \caption{Qualitative comparison of foreground segmentation results across different adaptation methods. GT denotes the ground truth foreground masks. The results illustrate the consistent effectiveness of DSGA \& LoRA in handling complex field conditions with varying occlusion, object scales and density.}
    \label{fig:qualitative_forground_evaluation}
\end{figure*}

\noindent Fig. \ref{fig:qualitative_forground_evaluation} illustrates a comprehensive visual comparison of foreground segmentation results across different adaptation approaches, comparing the proposed DSGA \& LoRA against baseline SAM, decoder-only fine-tuning, and other SOTA adaptation methods across representative field images. 

Qualitatively, DSGA \& LoRA demonstrated overall optimal performance across various scenarios, consistently capturing both prominent and occluded instances. It particularly excelled in complex field environments, producing accurate and coherent segmentation masks where other approaches failed to. Such robust performance was especially evident in rows 4, 5, 8 and 11, where it maintained precise object boundaries with the least misidentified foreground pixels, despite dense overlapping vegetation and multi-scale pods. 

Without adaptation, the original SAM model failed to distinguish foreground objects from complex backgrounds, resulting in overly conservative or completely blank segmentation. While mask decoder-only fine-tuning showed improvements over the original SAM baseline, it struggled with consistent object detection and misidentified similar components in the background, particularly in scenes with dense object distribution, as shown in rows 4, 5, 7, 9, and 10. 

Alternative methods, including BA, LoRA, and VPT variants, demonstrated varying degrees of success but failed to show comparable consistency as DSGA \& LoRA. BDL, while achieving competitive performance in moderate conditions, exhibited less stable results in extremely dense parts and complex background textures. Combined approaches, BA \& LoRA and BA \& DSGA, sometimes improved upon their individual components but still fell short for overall segmentation accuracy and consistency. These qualitative observations align with the quantitative metrics shown in Table \ref{tab:salient_sota_comparsion}. 

\begin{figure*}[htbp]
    \centering
    \includegraphics[width=\textwidth]{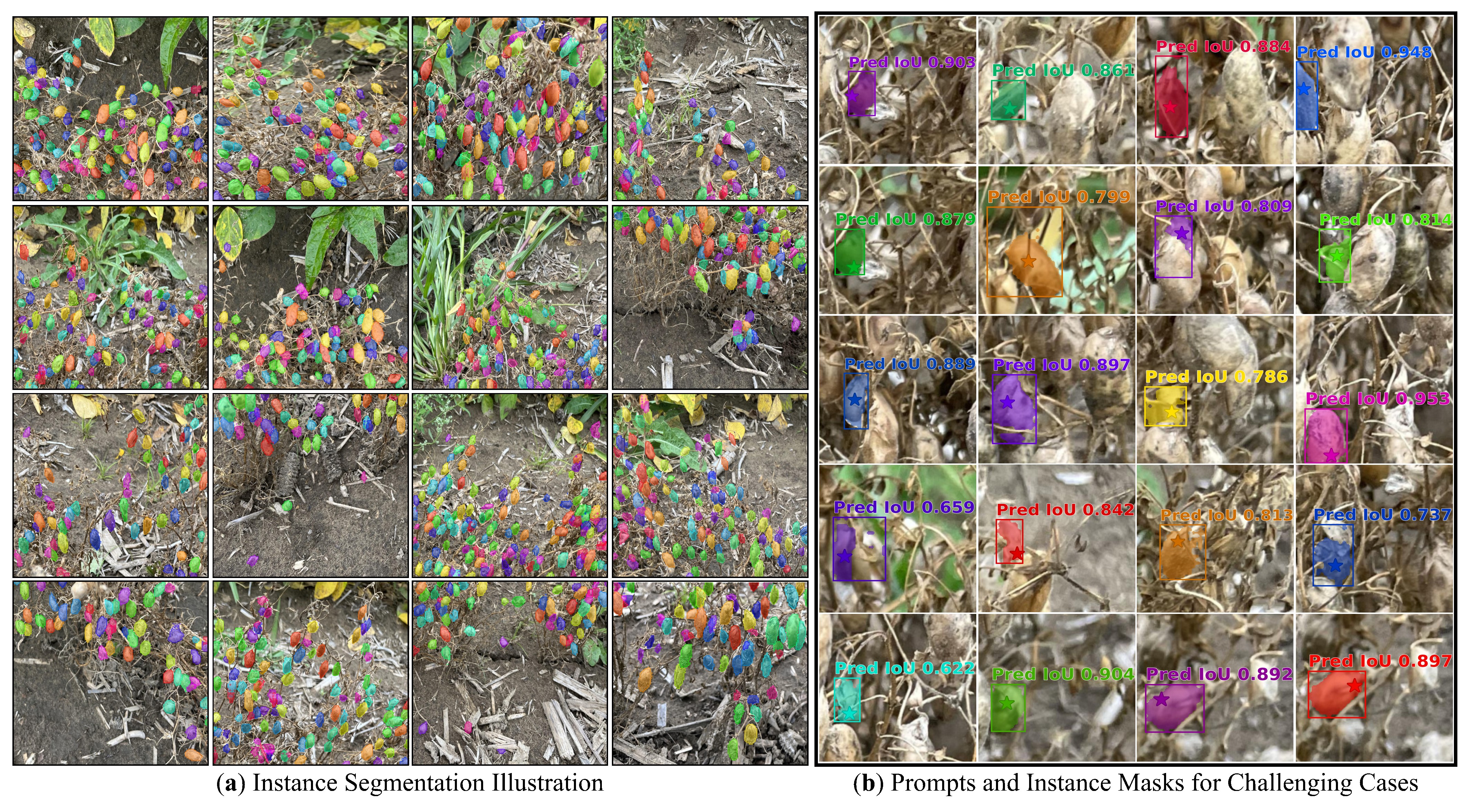}
    \caption{Qualitative illustration of instance segmentation results and challenging cases. (a) Instance segmentation of complex background instances. The model accurately captures the majority of the pods prompted. (b) Challenging instance segmentation cases, that mainly resulted from the high occlusion and incomplete capture of pods.}
    \label{fig:qualitative_instance_evaluation}
\end{figure*}

Fig. \ref{fig:qualitative_instance_evaluation} illustrates the instance segmentation results produced by DSGA \& LoRA, as well as challenging cases encountered. Fig. \ref{fig:qualitative_instance_evaluation}(a) depicts the model's ability to accurately identify instances in a complicated scene with varying numbers and scales of chickpea pods. Despite a few missing and misidentified instances, the majority of chickpea pod instances in the canopy were successfully detected. Fig. \ref{fig:qualitative_instance_evaluation}(b) exemplifies challenging cases that prove the method's adaptability to identify individual pods under intricate conditions. Overall, DSGA \& LoRA effectively segmented out partially or even mostly occluded pods.

\subsection{Ablation Studies}

\noindent Two sets of ablation experiments were conducted to evaluate the effectiveness and contribution of each key component in the proposed DSGA \& LoRA method, as well as the components in the proposed combined loss function. All ablation experiments utilized foreground segmentation results as the primary evaluation metric, as it represents the fundamental segmentation challenges independent of prompt-based guidance and directly impacts the subsequent instance-level segmentation performance.

\subsection{Analysis of DSGA Components}
\noindent The architectural evaluation experiments isolated and assessed the impact of each key component in DSGA and their integration with LoRA. Table \ref{tab: Abalation_on_DSGA} presents the component-wise performance contributions for different combinations of components, including Dynamic Graph Construction (DG) and Adaptive Pooling (AP), along with their integration with LoRA. The experiments also ablated two important configurations within DG, which are Learnable Weights (LW) and Polynomial Decay (PD) initialization.

The ablation results highlighted DG as the fundamental component for adaptation effectiveness. When implemented in isolation, DG without LW or PD achieved an $S_{\alpha}$ of 0.8271 and competitive $F_{\beta}$ and $E_{\xi}$ measures. This performance validated the effectiveness of similarity-based dynamic feature relationship modeling over static graph structures. 

Further analysis revealed important complementary effects between various architectural components. Applying PD initialization within DG further enhanced performance over most tested evaluation metrics. These results demonstrated that the non-linear decay pattern of PD initialization leads to improved feature representation by effectively emphasizing top-ranked neighbors more aggressively than linear schemes. LW enables dynamic weight adaptation of feature importance rankings during training, providing additional model flexibility for capturing complex feature relationships. As a result, LW increased the performance of DG across multiple metrics. Although AP did not produce very strong performance individually, AP improved the performance when integrated with different configurations of PD by further enhanced through improved local feature aggregation.

The integration of LoRA generally yielded consistent performance improvements across multiple evaluation metrics, demonstrating its complementary nature to graph-based feature adaptation. The only exception occurred when LoRA integrated with AP and DG without LW and PD, potentially due to the optimization conflicts of these two configurations. The full module achieved optimal performance across the evaluation metrics.

\begin{table*}[!htbp]
\centering
\caption{Performance contribution of each key component in the proposed DSGA \& LoRA for foreground segmentation. Top three results are highlighted in \textcolor{red}{\textbf{red}}, \textcolor{blue}{\textbf{blue}} and \textcolor{green}{\textbf{green}}, respectively.}
\setlength{\tabcolsep}{3pt}
\begin{tabular}{cccc|c|cccccccc}
    \hline
    \multicolumn{4}{c|}{\textbf{DSGA}}& \textbf{LoRA} & $S_{\alpha} \uparrow$ & $F_\beta^{\text{mean}} \uparrow$ & $F_\beta^{\text{max}} \uparrow$ & $F_\beta^{\text{adp}} \uparrow$ & $E_\xi^{\text{mean}} \uparrow$ & $E_\xi^{\text{max}} \uparrow$  &$E_\xi^{\text{adp}} \uparrow$&$\mathcal{M} \downarrow$  \\
    DG& AP& LW&PD& & &&&&& &&\\
    \hline
    \checkmark & &  && & 0.8271&0.8295&0.8661&0.8252&0.9472& 0.9606&0.9530&0.0406\\
     & \checkmark & & & & 0.8163& 0.8093& 0.8580& 0.7941& 0.9433& 0.9605 & 0.9466& 0.0430\\
     \checkmark & & & \checkmark & & 0.8322& 0.8149& \textcolor{green}{\textbf{0.8926}} & 0.7719& 0.9431& 0.9637& 0.9435& 0.0433\\
    \checkmark & \checkmark &  && & 0.8353&0.8236&0.8640&0.8126&\textcolor{blue}{\textbf{0.9509}}& \textcolor{green}{\textbf{0.9639}}&0.9551&\textcolor{green}{\textbf{0.0400}}\\
     \checkmark & \checkmark & & \checkmark & & 0.8235& 0.8238& 0.8903& 0.7825& 0.9327& 0.9628& 0.9441& 0.0451\\
     \checkmark & & \checkmark & & & 0.8351& \textcolor{green}{\textbf{0.8480}}& 0.8806& \textcolor{green}{\textbf{0.8403}}& 0.9469& 0.9621& \textcolor{green}{\textbf{0.9548}} & 0.0394\\
     \checkmark & & \checkmark & \checkmark & & 0.8234& 0.7917& 0.8871& 0.7274& 0.9313& 0.9623& 0.9216& 0.0478\\
     \checkmark & \checkmark & \checkmark  & & & 0.8210& 0.8239& \textcolor{blue}{\textbf{0.8935}}& 0.7747& 0.9284& 0.9624& 0.9382& 0.0466\\
     \checkmark & \checkmark & \checkmark  & \checkmark  & & 0.8305& 0.8165& 0.8867& 0.7779& 0.9385& 0.9633& 0.9449& 0.0442\\
    \hline
    \checkmark & &  && \checkmark & 0.8312&0.8224&0.8901&0.7880&0.9389& 0.9635&0.9442&0.0435\\
    & \checkmark & && \checkmark & 0.8287&0.8437&0.8821&0.8337&0.9415& 0.9577&0.9505&0.0410\\
     \checkmark & & & \checkmark & \checkmark & 0.8324& 0.8157& \textcolor{red}{\textbf{0.8949}} & 0.7719& 0.9377& \textcolor{blue}{\textbf{0.9641}}& 0.9391& 0.0440\\
    \checkmark & \checkmark & & & \checkmark & 0.8305& 0.8146& 0.8857& 0.7798& 0.9420& 0.9637& 0.9446& 0.0431\\
    \checkmark & \checkmark & & \checkmark  & \checkmark  & 0.8329& 0.8372& 0.8789& 0.8247& 0.9419& 0.9612& 0.9509& 0.0406\\
    \checkmark & &  \checkmark  && \checkmark & 0.8314&0.8017&0.8891&0.7573&0.9410& \textcolor{green}{\textbf{0.9639}}&0.9369&0.0441\\
     \checkmark & & \checkmark & & \checkmark & 0.8312& 0.8224& 0.8901& 0.7880& 0.9389& 0.9635& 0.9442& 0.0435\\
    \checkmark & &  \checkmark &\checkmark & \checkmark & \textcolor{blue}{\textbf{0.8402}} & \textcolor{red}{\textbf{0.8505}} &0.8869&\textcolor{blue}{\textbf{0.8412}}&\textcolor{green}{\textbf{0.9480}}& \textcolor{green}{\textbf{0.9639}}&\textcolor{blue}{\textbf{0.9565}}&\textcolor{blue}{\textbf{0.0390}}\\
    \checkmark & \checkmark & \checkmark  & & \checkmark & \textcolor{green}{\textbf{0.8380}}& 0.8401& 0.8846& 0.8337& 0.9446& 0.9637&0.9538&\textcolor{green}{\textbf{0.0400}}\\
    \checkmark & \checkmark & \checkmark  &\checkmark  & \checkmark & \textcolor{red}{\textbf{0.8453}} & \textcolor{blue}{\textbf{0.8495}} &0.8821& \textcolor{red}{\textbf{0.8428}} & \textcolor{red}{\textbf{0.9518}} & \textcolor{red}{\textbf{0.9650}} & \textcolor{red}{\textbf{0.9602}} & \textcolor{red}{\textbf{0.0384}}\\
    \hline
\end{tabular}
\label{tab: Abalation_on_DSGA}
\end{table*}

\subsection{Loss Function Components}

\noindent The ablation experiments for loss function analysis examined various combinations and weightings of the constituent loss terms to determine their individual and complementary effects, as shown in Table \ref{tab:Abalation_on_Loss_Func}. The baseline SAM architecture \cite{kirillov2023segment} employs a linear combination of Focal Loss and Dice Loss as $20\times \text{Focal Loss} + \text{Dice Loss}$. While this configuration demonstrated strong overall performance, with $S_{\alpha}$ of 0.8396 and $\mathcal{M}$ of 0.0440, it was observed that it frequently resulted in misclassifications of background components as foreground. This limitation significantly constrains the model's effectiveness with precise segmentation in fine-grained foreground segmentation tasks, especially for subtle boundary discrimination. 

To address these limitations, Binary Cross Entropy (BCE) loss was evaluated as an alternative baseline due to its established effectiveness in binary segmentation tasks. BCE demonstrated competitive performance in isolation, with $F_{\beta}^{\text{max}}$ of 0.9007 and $E_{\xi}^{\text{max}}$ of  0.9645, effectively addressing the foreground-background separation. However, the boundary misclassifications persisted, motivating the incorporation of Boundary Loss for enhanced edge discrimination.

The integration of Boundary Loss with the SAM's loss terms, Focal Loss and Dice Loss, brought significant performance improvements. While Boundary Loss alone showed limited effectiveness, its combination with Focal Loss and Dice Loss at unit scale (1:1:1) achieved optimal results across multiple metrics. This improvement can be attributed to Boundary Loss's complementary effect in enhancing boundary discrimination, working in concert with Dice Loss's region-based supervision and Focal Loss's class-balanced learning.

The study tried to optimize the weighting between these loss terms via comprehensive empirical testing and automatic weight adjustment through EMA. Contrary to theoretical expectations, the EMA-based dynamic weight adaptation demonstrated marginally decreased performance compared to the static weight configuration. This empirical finding suggested that while adaptive weight balancing potentially offers theoretical advantages in handling training dynamics, the stability of fixed weighting schemes may provide more consistent gradient flows for the optimization process in this specific segmentation task. Additional empirical weight testings with various loss weights demonstrated suboptimal performance across evaluation metrics when compared to the unit scale. These empirical results suggested that the balanced loss term weighting of 1:1:1 for the three loss terms provides optimal stability and accuracy for the proposed segmentation tasks.

\begin{table*}[!htbp]
\centering
\captionsetup{justification=raggedright,singlelinecheck=false}
\caption{Performance analysis of loss functions and EMA. Top three results are highlighted in \textcolor{red}{\textbf{red}}, \textcolor{blue}{\textbf{blue}} and \textcolor{green}{\textbf{green}}, respectively. FL: Focal Loss, DL: Dice Loss, BL: Boundary Loss, BCE: Binary Cross Entropy. }
\label{tab:Abalation_on_Loss_Func}
\small
\setlength{\tabcolsep}{3.5pt}
\begin{tabular}{cccc|c|cccccccc}
    \hline
    \multicolumn{4}{c|}{Loss Function} & EMA & $S_{\alpha} \uparrow$ & $F_\beta^{\text{mean}} \uparrow$ & $F_\beta^{\text{max}} \uparrow$ & $F_\beta^{\text{adp}} \uparrow$ & $E_\xi^{\text{mean}} \uparrow$ & $E_\xi^{\text{max}} \uparrow$ & $E_\xi^{\text{adp}} \uparrow$ & $\mathcal{M} \downarrow$ \\
    FL & DL & BL & BCE & & & & & & & & & \\
    \hline
    1 & & & & & 0.8240 & 0.7966 & \textcolor{green}{\textbf{0.8865}}& 0.7119 & 0.9178 & 0.9619 & 0.9106 & 0.0508\\
    & 1 & & & & 0.8253 & 0.7948 & 0.8331 & 0.7863 & 0.9417 & 0.9569 & 0.9442 & 0.0436\\
    & & 1 & & & 0.5687 & 0.4550 & 0.6477 & 0.4657 & 0.5012 & 0.9124 & 0.7619 & 0.1156\\
    & & & 1 & & 0.8355& 0.8113 & \textcolor{blue}{\textbf{0.9007}} & 0.7593 & 0.9372 & 0.9645 & 0.9349 & 0.0442\\
    1 & 1 & & & & 0.8334 & 0.8389& 0.8683 & \textcolor{blue}{\textbf{0.8335}}& \textcolor{green}{\textbf{0.9499}} & 0.9617 & \textcolor{green}{\textbf{0.9571}} & \textcolor{green}{\textbf{0.0405}}\\
    1 & & 1 & & & 0.8246 & 0.7883 & 0.8842 & 0.7094 & 0.9272 & 0.9624 & 0.9105 & 0.0492\\
    & 1 & 1 & & & 0.8214 & 0.8153 & 0.8712 & 0.7932 & 0.9428 & 0.9621 & 0.9468 & 0.0430\\
    1 & 1 & 1 & & & \textcolor{red}{\textbf{0.8453}} & \textcolor{red}{\textbf{0.8495}} & 0.8821 & \textcolor{red}{\textbf{0.8428}} & \textcolor{red}{\textbf{0.9518}} & \textcolor{red}{\textbf{0.9650}} & \textcolor{red}{\textbf{0.9602}} & \textcolor{red}{\textbf{0.0384}}\\
    1 & 1 & 1 & & \checkmark & \textcolor{blue}{\textbf{0.8428}} & \textcolor{blue}{\textbf{0.8419}} & 0.8829 & \textcolor{green}{\textbf{0.8307}} & \textcolor{blue}{\textbf{0.9508}} & \textcolor{blue}{\textbf{0.9648}} & \textcolor{blue}{\textbf{0.9578}} & \textcolor{blue}{\textbf{0.0391}}\\
    20 & 1 & & & & \textcolor{green}{\textbf{0.8396}} & \textcolor{green}{\textbf{0.8450}} & \textcolor{red}{\textbf{0.9104}} & 0.7895& 0.9254 & \textcolor{green}{\textbf{0.9646}} & 0.9480 & 0.0440\\
    20 & 1 & 1 & & & 0.8302 & 0.7934 & 0.8795 & 0.7399 & 0.9313 & 0.9611 & 0.9278 & 0.0477\\
    20 & 1 & 1 & & \checkmark & 0.8252 & 0.7947 & 0.8806 & 0.7461 & 0.9333 & 0.9617 & 0.9291 & 0.0467\\
    \hline
\end{tabular}
\end{table*}

\subsection{Model Interpretation}

\noindent Two complementary visualization methods were employed to better visualize and interpret the result of the proposed adaptation modules, which were Gradient-weights Class Activation Mapping (Grad-CAM) \cite{selvaraju2017grad} and t-distributed Stochastic Neighbor Embedding (t-SNE) \cite{van2008visualizing}.

\subsubsection{Attention Visualization vis Grad-CAM}
\begin{figure*}[htbp]
    \centering
    \includegraphics[width=\linewidth]{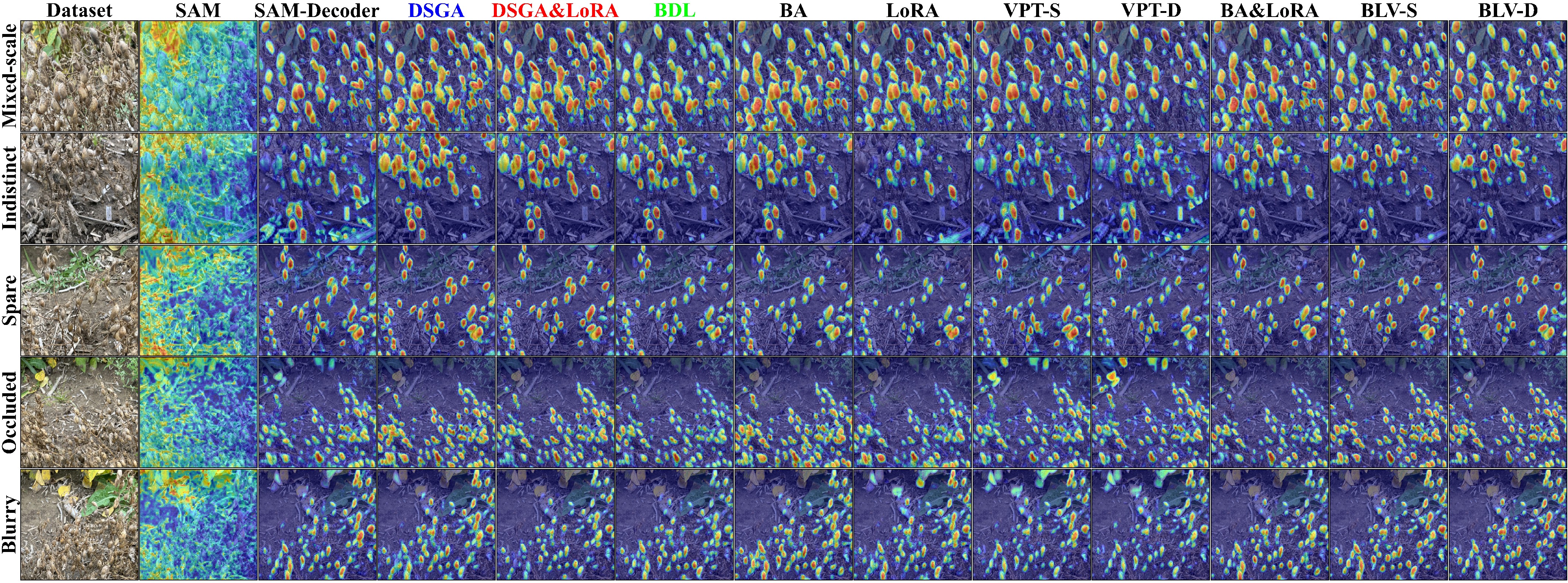}
    \caption{Grad-CAM attention visualization analysis for challenging segmentation scenarios. Visualization maps from different adaptation variants are presented for five representative challenging cases: mixed-scale environments, indistinct conditions, sparse distributions, occluded scenes, and blurry conditions. Warmer colors (yellow to red) indicate regions of higher attention importance while cooler colors (blue to green) represent areas of lower importance.}
    \label{fig:grad-cam}
\end{figure*}

\noindent Fig. \ref{fig:grad-cam} presents Grad-CAM visualizations across different adaptation variants for five representative challenging scenarios, which were mixed-scale environments, indistinct conditions, sparse distributions, occluded scenes, and blurry conditions. The visualization analysis at the decoder's output upscaling layer provided visual explanation to help interpret feature extraction ability of different adaptation methods during the final stage of mask generation.

The original SAM exhibited non-discriminative activation patterns in all scenarios, suggesting limited feature discrimination ability on the complex chickpea dataset. Overall, the proposed DSGA \& LoRA and DSGA demonstrated the most discriminative activation patterns, with warm colors highlighted pod regions against deep blue backgrounds. These distinct patterns indicated enhanced discrimination between foreground objects and complex backgrounds at the upscaling layer. While most variants showed different levels of improvements over the baseline, DSGA \& LoRA achieved the most consistent and well-defined feature localization. In the mixed-scale scenario, features produced by DSGA \& LoRA and DSGA maintained precise activation patterns regardless of object size variations, while these patterns were not captured in other adaptation methods. DSGA \& LoRA and DSGA also exhibited exceptional capability in all other tested complex scenes, robustly separating foreground pods from background and clearly delineating object boundaries. In contrast, other tested adaptation methods and their compositions showed inaccurate activation patterns and intensities, especially in the mixed-scale and indistinct scenes. 

Interestingly, DSGA alone produced more concentrated activation patterns in the sparse and occluded scenarios, whereas DSGA \& LoRA yielded more balanced and distributed feature localization across object regions. This difference suggested that interpreting LoRA's attention modification mechanism potentially modulates feature localization, trading off some activation intensity for improved contextual awareness and boundary precision. 

\subsubsection{Attention Visualization via t-SNE}

\begin{figure*}[htbp]
    \centering
    \includegraphics[width=0.85\textwidth]{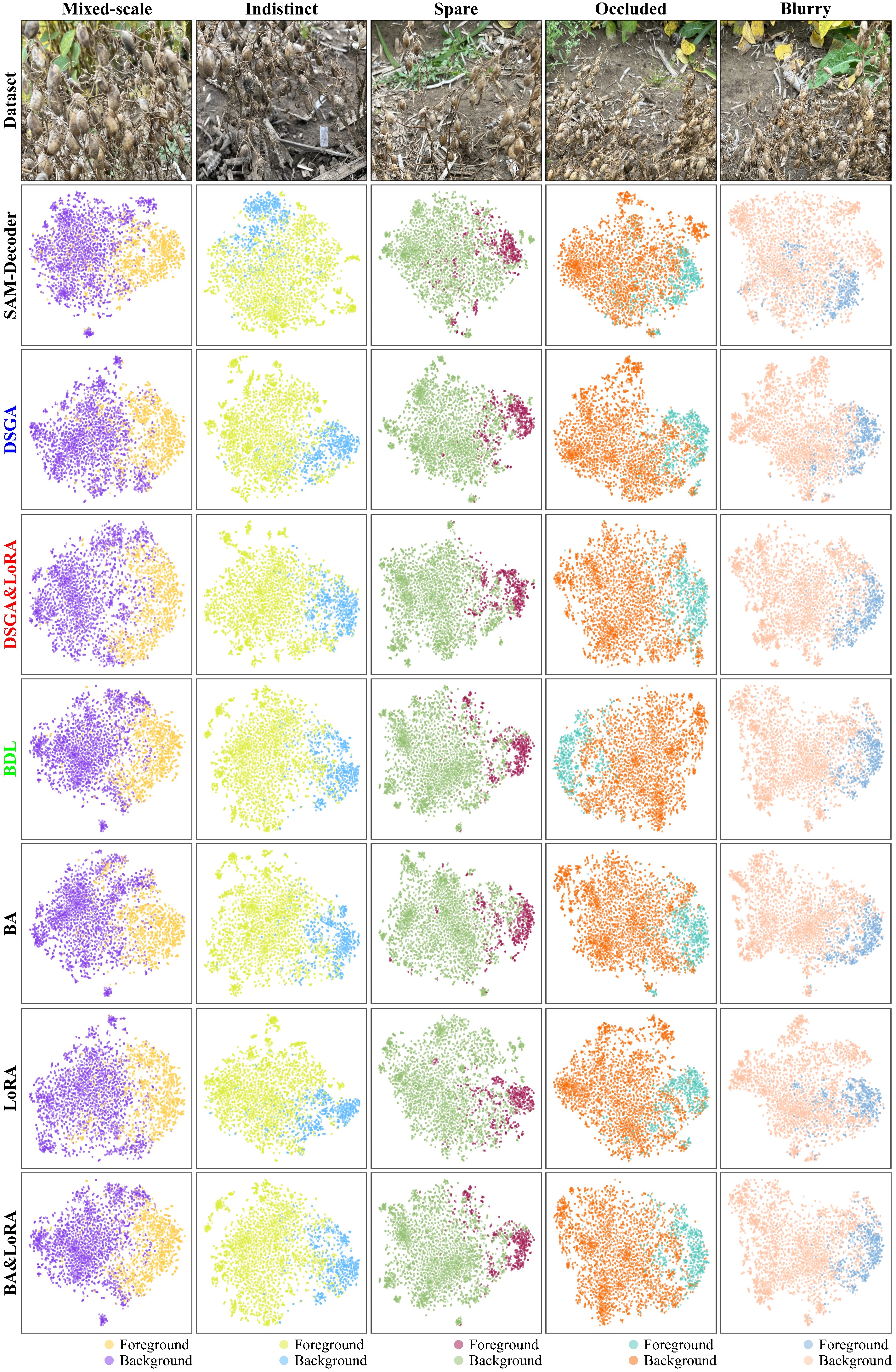}
    \caption{t-SNE visualization of feature embeddings across different adaptation variants and challenging scenarios. The visualization projects the high-dimensional 768D feature space into 2D, comparing feature distributions for six overall best-performing adaptation variants and the baseline model under five challenging scenarios. Each point in the t-SNE projection represents a locality-specific feature embedding. DSGA \& LoRA achieves the overall most distinct cluster separation across all scenarios, particularly in occluded and blurry conditions.}
    \label{fig:t-SNE}
\end{figure*}

\noindent Fig. \ref{fig:t-SNE} presents t-SNE visualizations of the 768-dimensional (768D) image embeddings projected into 2-dimensional (2D) space. The visualizations helped interpret the learned feature representations across different challenging scenarios via distinctly clustered patterns of foreground and background feature embeddings. Due to space constraints, only the t-SNE visualizations of the baseline model with SAM mask decoder fine-tuned and the six best-performing adaptation methods were presented.

DSGA \& LoRA demonstrated well-defined feature separation in most scenarios. While clusters of other adaptation methods produced severe overlapping, boundaries were generally clearer between foreground and background embeddings generated by DSGA \& LoRA. Feature overlappings were observed in all visualizations in the indistinct condition, while DSGA \& LoRA exhibited better separation between foreground and background features with minor ambiguity in some boundary regions. The sparse distribution scenario also illustrated DSGA \& LoRA's capability to maintain coherent feature clusters despite limited spatial context. In occluded scenes, it also successfully preserved feature relationships between partially visible objects, as shown by the continuous and well-structured distribution of foreground clusters. In mixed-scale conditions, DSGA \& LoRA generated clearly separated embedding clusters with features of varying scales. The blurry condition analysis also demonstrated robust feature extraction of DSGA \& LoRA, despite of degraded image quality.

However, several limitations of DSGA \& LoRA were apparent in the visualization. While it achieved better overall clustering, misclassified or overlapping embeddings consistently existed in all tested scenarios.  Less compact clustering was also observed in highly occluded scenarios, compared to BA \& LoRA. These inaccurate clustering patterns indicated potential enhancement directions for DSGA \& LoRA to handle ambiguous features and potential inconsistent performance in occlusion regions. These visualization interpretations collectively demonstrated DSGA \& LoRA's enhanced feature representation capability over other tested methods, while highlighting potential improvement when handling extremely challenging conditions.

\subsection{Field Organ Instance Counting}
\begin{figure*}[htbp]
    \centering
    \includegraphics[width=\textwidth]{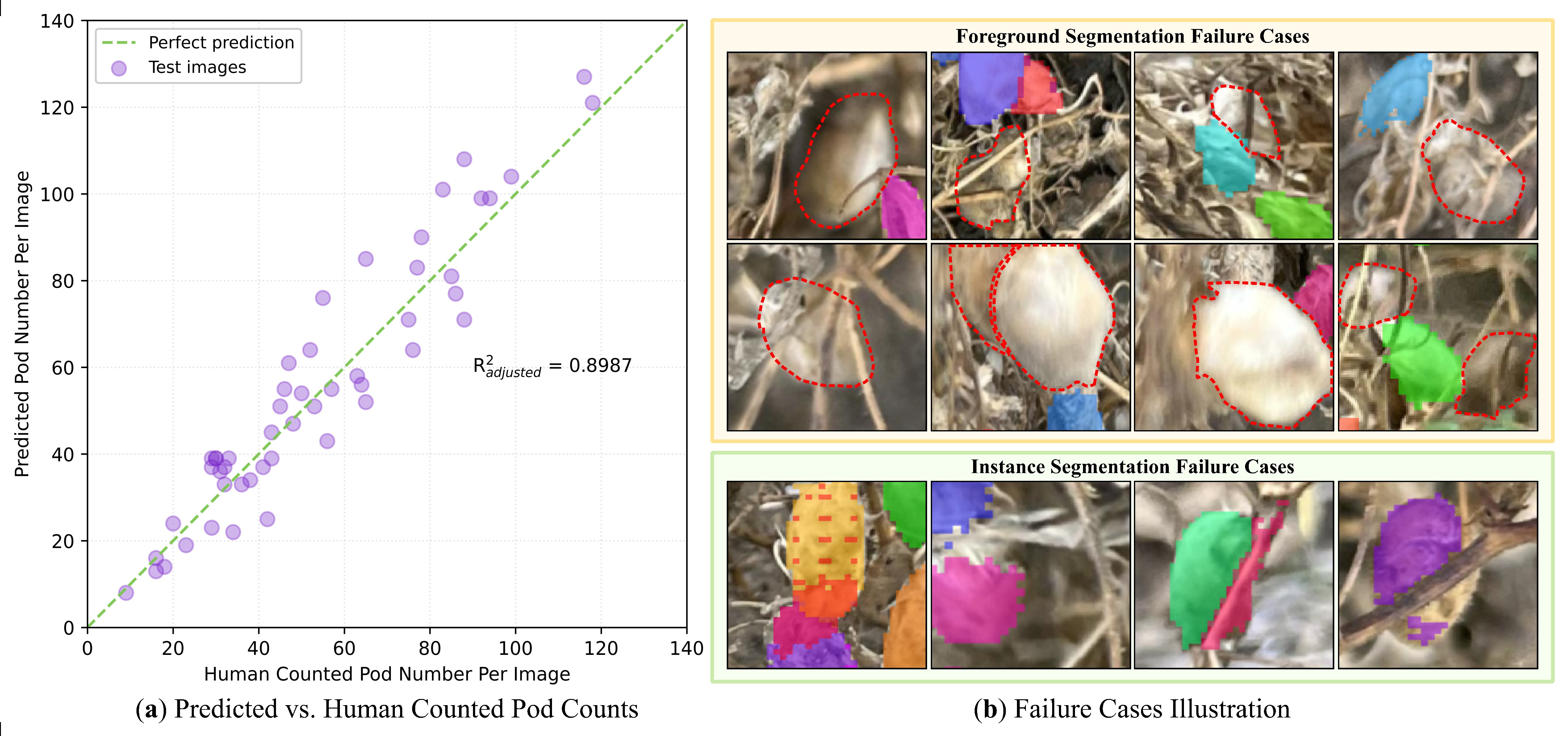}
    \caption{Quantitative evaluation of the proposed model's performance on chickpea pod counting and current detection limitations. (a) Scatter plot illustrating the correlation between predicted pod instance counts and manually annotated ground truth values across the test dataset. The green line represents the ideal prediction. (b) Representative failure cases demonstrating detection limitations under challenging field conditions. Regions with motion blur artifacts, extremely severe occlusion, and low foreground-background contrast yielded reduced detection performance during foreground segmentation, consequently affecting instance-level detection.}
    \label{fig:field_performance}
\end{figure*}

\noindent The proposed framework was evaluated in real-world chickpea field conditions to assess its practical effectiveness for automated organ instance counting applications. Fig. \ref{fig:field_performance}(a) presents a comprehensive analysis of the model's field counting results, demonstrating a strong positive correlation between predicted and human-counted pod numbers across diverse field conditions. The scatter plot demonstrated that most data points clustered near the perfect prediction line with an adjusted $\text{R}^2$ value of 0.8987, indicating accurate counting in most tested cases. Overall, the model exhibited consistent performance across various field conditions, with pod densities ranging from approximately 10 to 120 pods per image. A slight decline in accuracy appeared in high-density occluded scenarios.

To investigate these inaccuracies, Fig. \ref{fig:field_performance}(b)  delineates representative error cases that constrained the model's performance, classified into two primary categories. During foreground segmentation, the model struggled to identify extremely blurry or heavily occluded pods. These detection failures typically stem from motion blur or camera focus issues, resulting in inadequate saliency and, consequently, no prompt point generation for these pods. With respect to instance segmentation, structural instance segmentation errors arise from extremely complex occlusion patterns where pods intersected with branches, leaves, or other pods, creating ambiguous boundaries or fragmenting single-pod instances into discrete pixel groups. This limitation is pronounced in densely clustered regions where boundary definition became problematic and multiple overlapping pods created complex visual patterns that challenged the model's instance discrimination capabilities.

\subsection{Limitations and Future Directions}
\noindent The experimental analysis reveals several practical limitations in real-world deployment scenarios. As illustrated in Fig. \ref{fig:field_performance}, the model's performance degrades in extremely dense canopy conditions due to occlusion and overlapping pods, while blurry and low-contrast features further impact segmentation reliability. These challenges affect the system's applicability in variable field conditions where accurate, consistent segmentation is crucial for precise crop monitoring. 

Addressing these limitations requires consideration of both architectural constraints and potential enhancements. A fundamental architectural constraint stems from the SAM backbone's fixed 1024 × 1024 input resolution, which limits the model from capturing more fine-grained details essential for precise segmentation of small plant organs. To address similar challenges, recent work by \cite{ke2023segment} inserts learnable output tokens to refine the SAM mask decoder to generate high-resolution masks, representing a complementary approach to address this limitation. While the present study prioritizes the adaptation of the image encoder, these studies focus on adapting SAM's prompt encoder and mask decoder, representing a distinct but complementary research direction \cite{wu2025medical, ke2023segment}.

Beyond resolution limitations, current model backbone choices also present essential computational efficiency-performance trade-offs. The superior segmentation performance of the proposed framework derives not only from the efficacy of the adaptation mechanisms but also fundamentally from the representational capacity of the preserved parameters of pre-trained SAM \cite{aghajanyan2020intrinsic}. \cite{kirillov2023segment} demonstrates a more advanced performance of SAM with ViT-Large and ViT-Huge backbones. Shifting from ViT-Base to ViT-Large and ViT-Huge will likely increase the segmentation performance. However, such an upgrade will also increase the frozen model parameters and subsequently increase the learnable parameters in adaptation modules, resulting in higher computational requirements. Considering model capacity and computational efficiency, the current implementation of this study still utilizes the ViT-Base backbone. A future direction might address these trade-offs through model pruning and optimization, improving practical applicability for large-scale field deployments \cite{zhang2023faster}.

For future studies, architectural enhancements could potentially improve upon these limitations, such as integrating multi-view imaging systems and imagery with higher spectral resolution and additional modality to enhance the model's capacity to discriminate fine-grained morphological features under variable field conditions  \cite{li2024light, song2024multispectral, wang2024adapting}.

While this study focuses specifically on chickpea pod segmentation as a case study. Beyond addressing the limitations discussed above, future research should explore expanded applications of the system. Such as to adapt the vision foundation model toward greater generalizability for cross-stage and cross-crop adaptation studies. Additionally, current 2D segmentation can be projected onto 3D feature space for enhanced spatial understanding, enabling more comprehensive analysis of complex 3D structures and their relationships across multiple dimensions \cite{yang2023sam3d}. These extensions would enhance the system's utility across diverse agricultural applications.

\section{Conclusion}
\noindent This study presents a two-stage framework for few-shot foreground and instance segmentation of small dense plant organs in complex agricultural backgrounds. The proposed module, DSGA, synergistically integrates with LoRA to achieve PEFT on top of SAM with up to 10 training samples. This lightweight composition comprises only 4.33M trainable parameters, which is 4.62\% of the original SAM, and requires 32.69 GFLOPs of computation, demonstrating significant computational efficiency while maintaining advanced feature extraction capabilities.

Comprehensive experiments validate and interpret the proposed framework through rigorous comparison with SOTA modules and their integrations. The DSGA module effectively captures both global and local feature dependencies through its dynamic similarity adjacency graph construction mechanism. Its integration with LoRA enables efficient parameter updates within the image encoder's attention layers. This synergistic composition results in significant quantitative improvements, demonstrating superior performance on both foreground and instance segmentation tasks across different tested shots. Ablation studies identify the effectiveness of proposed core components in DSGA and the integration of LoRA, while establishing that a balanced combination of Focal, Dice, and Boundary Losses exhibits the capability to distinguish small dense objects with ambiguous boundaries in heterogeneous field conditions. 

Current technical limitations persist in the input resolution of SAM and segmentation performance degradation in challenging scenarios, especially for objects in extremely blurry and heavily occluded conditions. Despite significantly improved parameter efficiency, the computational demand might still prevent implementing this framework for more resource-constrained field deployments. Future research could extend this work through multi-view or multispectral imaging systems to mitigate extreme occlusion challenges and broader investigations into cross-stage and cross-crop adaptability of SAM and PEFT modules. 

\bibliographystyle{IEEEtran}
\bibliography{bibliography}

\end{document}